\documentclass[runningheads]{llncs}

\usepackage{eccv}
\usepackage{booktabs}

\usepackage{eccvabbrv}
\usepackage{graphicx}
\usepackage{booktabs}
\usepackage{multirow}
\usepackage[accsupp]{axessibility}  
\usepackage{bbding}
\usepackage{algpseudocode}
\usepackage{algorithm}
\usepackage{algcompatible}
\usepackage{orcidlink}
\begin{document}
	\title{Superpixel-informed Implicit Neural Representation for Multi-Dimensional Data} 
	\titlerunning{Superpixel-informed INR for Multi-Dimensional Data}
	
	\author{Jiayi Li\inst{1}\orcidlink{0000-0003-1626-4759} \and
		Xile Zhao\inst{1}\orcidlink{0000-0002-6540-946X} \and
		Jianli Wang\inst{2}\orcidlink{0000-0003-4774-4894} \and
		Chao Wang\inst{3}\orcidlink{0000-0001-6524-504X} \and
		Min Wang\inst{4}\orcidlink{0000-0003-3083-2521}}
	
	\authorrunning{J. Li et al.}
	
	\institute{University of Electronic Science and Technology of China, Chengdu, China
	\email{lijiayi03531@gmail.com, xlzhao122003@163.com}
	\and
	Southwest Jiaotong University, Chengdu, China\\
	\email{wangjianli$\_$123@163.com}
	\and
	Southern University of Science and Technology, Shenzhen, China\\
	\email{chaowang.hk@gmail.com}
	\and
	Jiangxi University of Finance and Economics, Nanchang, China\\
	\email{minwang1989@126.com}
}
	
	\maketitle
	\begin{abstract}
		Recently, implicit neural representations (INRs) have attracted increasing attention for multi-dimensional data recovery. However, INRs simply map coordinates via a multi-layer perceptron (MLP) to corresponding values, ignoring the inherent semantic information of the data. To leverage semantic priors from the data, we propose a novel Superpixel-informed INR (S-INR). Specifically, we suggest utilizing generalized superpixel instead of pixel as an alternative basic unit of INR for multi-dimensional data (e.g., images and weather data). The coordinates of generalized superpixels are first fed into exclusive attention-based MLPs, and then the intermediate results interact with a shared dictionary matrix. The elaborately designed modules in S-INR allow us to ingenuously exploit the semantic information within and across generalized superpixels. Extensive experiments on various applications validate the effectiveness and efficacy of our S-INR compared to state-of-the-art INR methods.
	\end{abstract}
	%
	\section{Introduction}
	
	Implicit neural representations (INRs)\cite{sitzmann2020implicit} have attracted significant research interest in recent years as a novel method for representing multi-dimensional data, such as images \cite{images1, images2, images3}, point clouds\cite{point1, point2}, videos\cite{vedio1, vedio2, vedio3}, and audio signals\cite{auto1, auto2}. INR utilizes a coordinate-based multilayer perceptron (MLP), which takes spatial coordinates as input and outputs corresponding values. Essentially, INR implicitly represents data via deep neural networks to achieve a continuous representation. The strong representation capacity of INR opens up new possibilities for multi-dimensional data modeling and analysis, overcoming the limitations of traditional discrete representations that rely on fixed mesh-grids or matrices.

	\begin{figure*}[ht]
		\centering
		\includegraphics[width=0.9\textwidth]{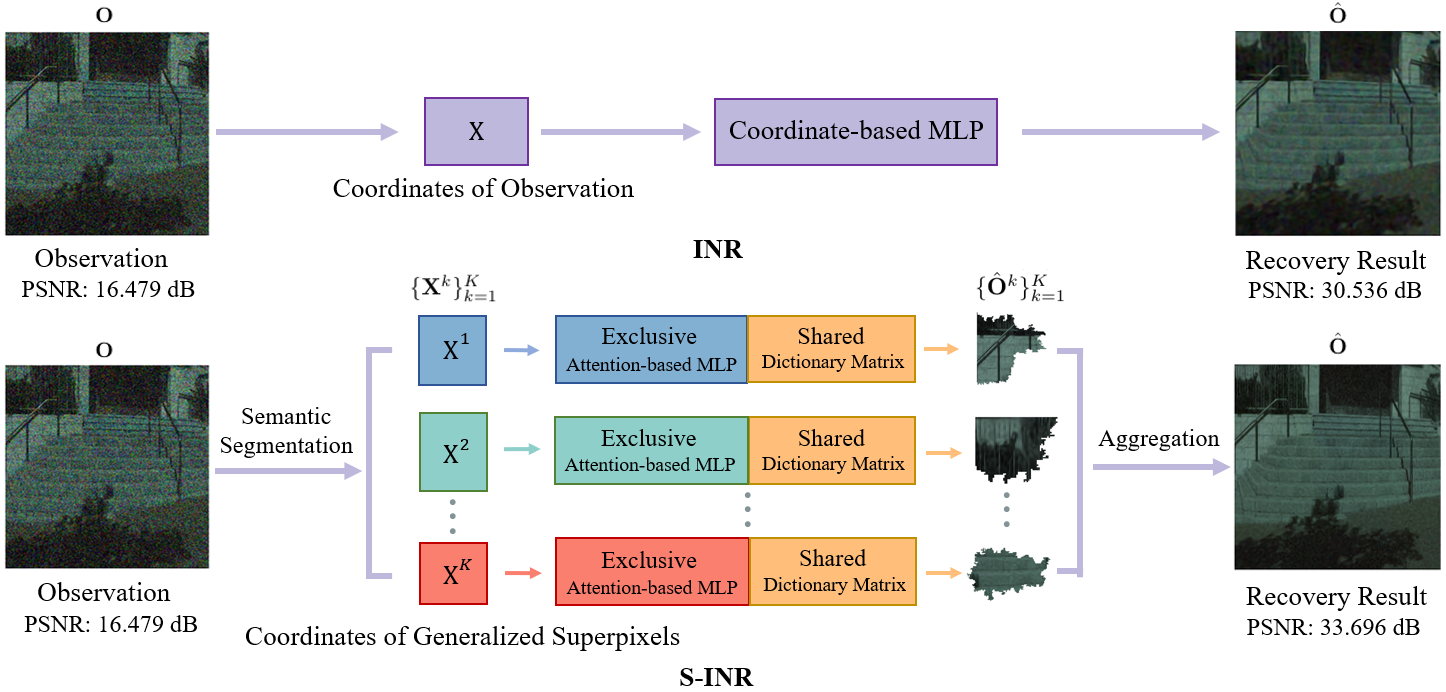}

		\caption{Illustration of the overall processing of the traditional INR and  our proposed Superpixel-informed INR (S-INR) on image denoising task. In the proposed S-INR, generalized superpixels are used as basic units. The coordinates of generalized superpixels are fed into the exclusive attention-based MLPs, and then the intermediate results interact with a shared dictionary matrix to obtain recovery results.}
		\label{fig:method}

	\end{figure*}

	INR has achieved success in a variety of applications on multi-dimensional data recovery tasks with its strong representation capacity, including image super-resolution\cite{image3, image4, 9991174}, completion\cite{inpainting1, inpainting2}, generation\cite{inverse1, inverse2}, and enhancement \cite{image1}, 3D shape representation\cite{t1, t2, t3}, novel view synthesis \cite{n1, n2, n3}, and physics-based simulation and inference\cite{simulation, Hofherr_2023_WACV}. Recently, INR, i.e., coordinate-based MLP, has been shown to perform poorly in encoding signals with high-frequency components, due to the corresponding neural tangent kernel (NTK) of MLPs being prone to high-frequency fall-offs \cite{nips_mlp_high_fre, pmlr_mlp_high_fre}. To address the above issue, researchers have focused on improving the expressiveness of the INR via two primary aspects: positional encoding and nonlinear activation functions. For positional encoding, the prevalent methods used Fourier features to project low-dimensional inputs into higher-dimensional space by applying a positional encoding layer before the MLP \cite{tancik2020fourfeat, nerf}. Later, Long et al. \cite{Mai_2022_CVPR} proposed a phase-varying positional encoding module that exploits the relationship between phase information in sinusoidal functions and their displacements. For nonlinear activation functions, Sitzmann et al. used a periodic activation function, Ramasinghe et al.\cite{eccv_activation} proposed a unified framework for activation, Saragadam et al. \cite{WIRE} used a continuous Gabor wavelet activation function, Shen et al. proposed \cite{shen2023trident} a novel function TRIDENT for INR characterized by a trilogy of nonlinearities, and Danzel et al.\cite{serrano2024hosc} introduced the hyperbolic oscillation function, which is designed to capture high-frequency information of the underlying data. Other improvement methods such as band-limited coordinate networks\cite{bacon} and multiplicative filter networks\cite{mfn} have also been proven effective.

	Although INR and its variants have achieved success in various applications\cite{miner, kilo, acorn, wa1, wa4}, previous INR methods ignore the rich inherent semantic information of the data, which could be beneficial for data representations and subsequent applications. A natural question followed: can we develop a novel data representation that efficiently exploits the inherent semantic information of the data under the INR framework?

	To address this challenge, we first suggest utilizing generalized superpixels instead of pixels as alternative basic units of INR for multi-dimensional data, which encode rich semantic information. The generalized superpixels, going beyond traditional superpixels, are not limited to image data, but can also be applied to point data arising from real-world applications in general. Then, to fully exploit the semantic information within and across generalized superpixels, we propose a novel Superpixel-informed INR (S-INR). Specifically, we elaborately design two key modules in S-INR, i.e., exclusive attention-based MLPs and a shared dictionary matrix. The exclusive attention-based MLP enhances the expressiveness of S-INR in feature dimensions within each generalized superpixel. The shared dictionary matrix allows us to capture correlations between generalized superpixels. Fig. \ref{fig:method} clearly illustrates the superior representation capacity of our proposed S-INR when compared to traditional INR. Our proposed S-INR effectively exploits the semantic information within and across generalized superpixels, resulting in outstanding performance. The contributions of this paper are as follows:

(\romannumeral1) We suggest utilizing generalized superpixels instead of pixels as alternative basic units of INR, which encode rich semantic information. The generalized superpixels are not limited to image data, but also suitable for more general point data arising from real-world applications than traditional superpixels.

(\romannumeral2) To exploit the semantic information within and across generalized superpixels, we propose a novel superpixel-informed implicit neural representation (termed as S-INR). The key modules in S-INR, i.e., exclusive attention-based MLPs and the shared dictionary matrix, are elaborately designed to respect the individuality of each generalized superpixel and capture the commonalities between them.

(\romannumeral3) Extensive experiments including image reconstruction, image completion, image denoising, weather data completion, and 3D surface completion, validate the broad applicability and superiority of S-INR compared to state-of-the-art INR methods.

\section{The Proposed Model}

\textbf{Notation.} $x,\;\textbf{x}, \text{and} \; \mathbf{X}$ denote the scalar, vector, and matrix, respectively. Here, $\|\cdot\|$ denotes the vector norm.

\subsection{Definition of Generalized Superpixel}

Pixels\footnote{For point data, pixels correspond to a feature value vector.} are considered as the basic unit of INR for multi-dimensional data. However, existing INRs, utilizing pixels as the basic unit, do not effectively exploit the inherent semantic information of the data. To address this problem, we propose a generalized superpixel and employ it as an alternative basic unit to enable the exploitation of the inherent semantic information of multi-dimensional data as priors under the INR framework.

\begin{definition}[Generalized Superpixel]
	Given a point data $\mathbf{O}\in\mathbb{R}^{s\times n}$ consisting of $n$ points, i.e., $\{{\textbf o_i}\in\mathbb{R}^{s}\}_{i=1}^n$, where $s$ is the number of feature dimensions, and the number of generalized superpixels $K$. $\{{\bf O}^{k}=\{{\mathbf o}_{i}^k\}_{i=1}^{n_k}\}_{k=1}^{K}$, $n_1+\cdots+n_{K}=n$, representing assigned groups of data points are called generalized superpixels if and only if they satisfy the following two conditions:
	\begin{enumerate}
		\item \textbf{Disjointness.} $\{{\bf O}^{k}\}_{k=1}^{K}$ are pairwise disjoint, i.e., $ {\bf O}^{k}\cap{\bf O}^{k'}=\emptyset$ for all $k \neq k'$.
		\item \textbf{Spatial Connectivity.} Each $\{{\bf O}^{k}\}_{k=1}^{K}$ forms a spatially connected blob, meaning that all data points within a generalized superpixel are contiguous or proximate in the spatial domain.
	\end{enumerate}	
\end{definition}

\begin{figure}[!htpb]
	\centering
	\subfloat[Image data]{\label{sp_i} \includegraphics [width=0.22\textwidth]{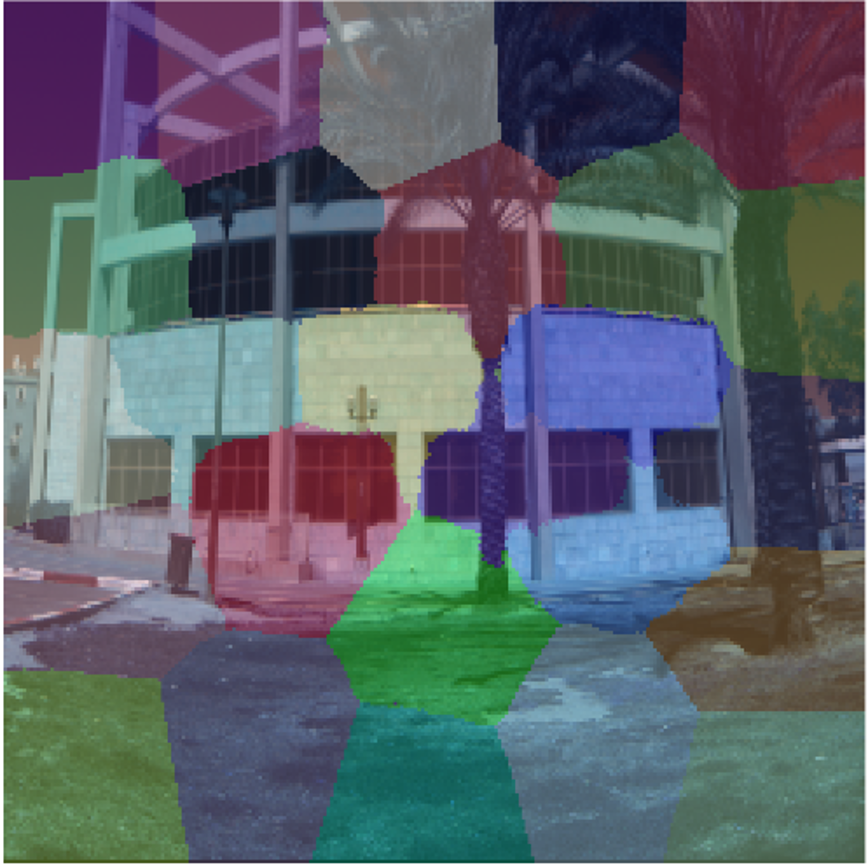}}
	\;\;\;\;\;\;
	\subfloat[3D surface data]{\label{sp_p} \includegraphics [width=0.25\textwidth]{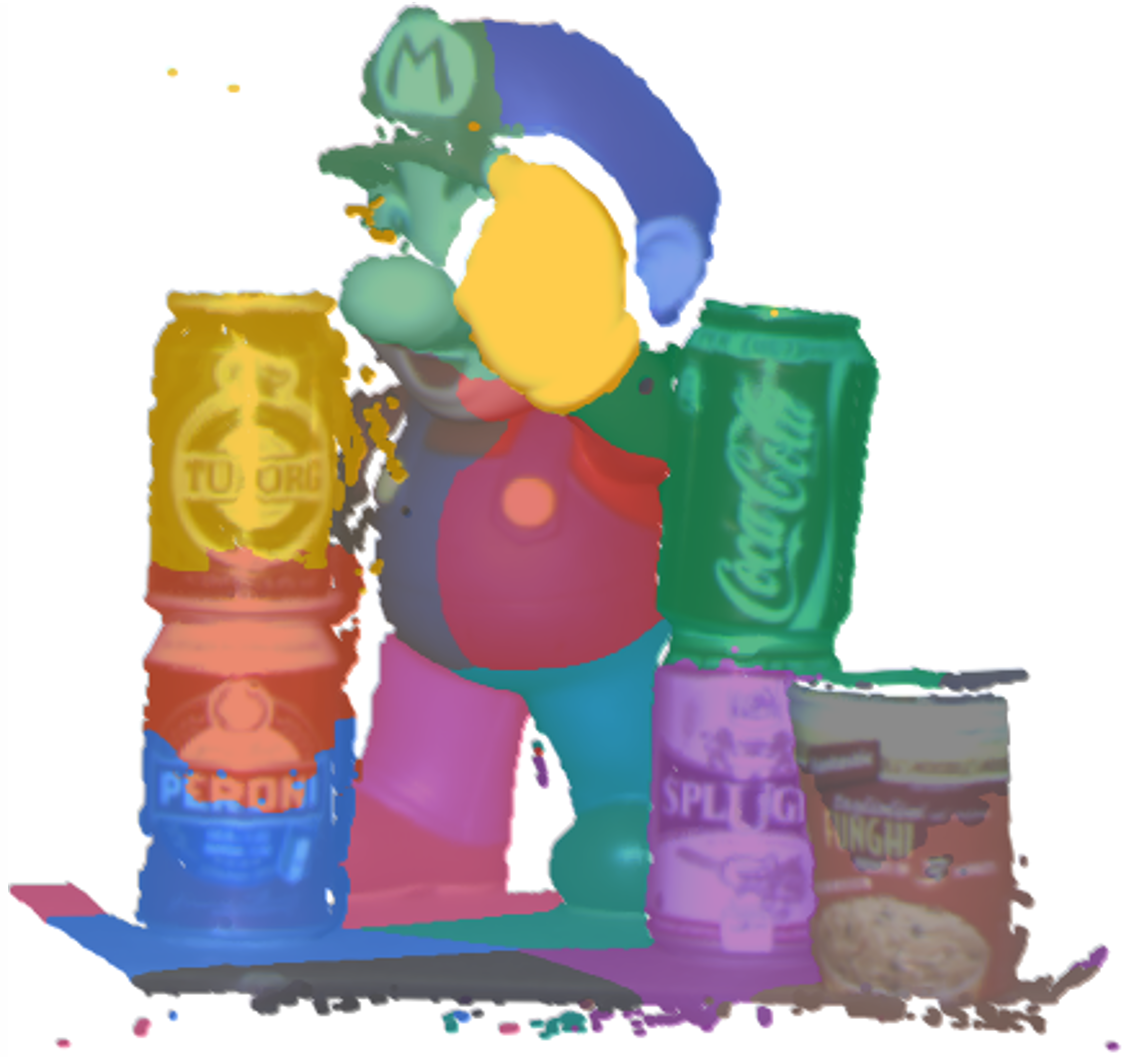}}
	\;\;\;\;
	\subfloat[Weather data]{\label{sp_w} \includegraphics [width=0.25\textwidth]{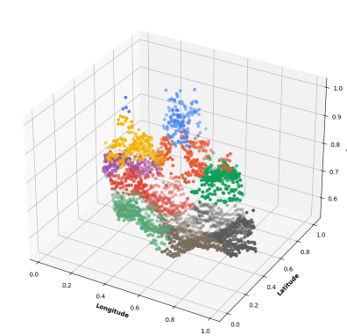}}
	
	\caption{Illustration of generalized superpixel segmentation on the (a) image data, (b) 3D surface data, and (c) weather data.}\label{superpixel}
\end{figure}

\begin{remark}
	The traditional superpixel is limited to image data. In contrast, our proposed generalized superpixel is beyond image data. It is also suitable for more general point data arising from real-world applications, such as 3D surface data and weather data, as illustrated in \cref{sp_p,sp_w}.	
\end{remark}

\begin{algorithm}[htbp!]
	\small
	\renewcommand{\algorithmicrequire}{\textbf{Input:}}
	\renewcommand{\algorithmicensure}{\textbf{Output:}}
	\caption{The Generalized Superpixel Segmentation Algorithm (GSSA)}
	\label{alg1}
	\begin{algorithmic}[1]
		\REQUIRE The point data $\{{\textbf o_i}\in\mathbb{R}^{s}\}_{i=1}^n$ and its coordinate $\{{\textbf x_i}\in\mathbb{R}^{c}\}_{i=1}^n$, weight $\alpha$, and the number of generalized superpixels $K$.
		
		\!\!\!\!\!\!\!\!\!\!\!\!\!\!\!\!\!\!\!\!\!\!\textbf{Initialization:}
		Set initial clustering centers $\boldsymbol{\mu}_1,\;\dots,\;\boldsymbol{\mu}_K$ using the $k$-means$\overset{\scriptstyle\mathrm{++}}{}$ initialization method and $\{{\mathbf x}_{\boldsymbol{\mu}_k}\}_{k=1}^{K}$ are their corresponding coordinates.
		\REPEAT
		\FOR{$i=1$ to $n$}
		\STATE $ m_{ik}=\begin{cases} 1 & \text{if } k = \arg\min_k \|\textbf{o}_i-\boldsymbol{\mu}_k\|^2 + \alpha\|\textbf{x}_i-{\mathbf x}_{\boldsymbol{\mu}_k}\|^2 \\ 0 & \text{otherwise } \end{cases} $
		\ENDFOR
		\FOR{$k=1$ to $K$}
		\STATE $ \; \boldsymbol{\mu}_k=\frac{\sum_{i=1}^{n} m_{ik}{\textbf o}_i}{\sum_{i=1}^{n} m_{ik}} $
		\STATE $ \mathbf{x}_{\boldsymbol{\mu}_k}=\frac{\sum_{i=1}^{n} m_{ik}{\textbf x}_i}{\sum_{i=1}^{n} m_{ik}} $
		\ENDFOR
		\UNTIL{Convergence}
		\ENSURE Generalized superpixels ${\mathbf O}^k=\{ {\mathbf o}_i |m_{ik} = 1, i=1,\;\dots,\;n \}$ and coordinates ${\mathbf X}^k=\{ {\mathbf x}_i |m_{ik} = 1, i=1,\;\dots,\;n \}$, $k=1,\;\dots,\;K$.
	\end{algorithmic}  
\end{algorithm}

Next, we will introduce our method for finding generalized superpixels from point data. General clustering algorithms (e.g., $k$-means$\overset{\scriptstyle\mathrm{++}}{}$\cite{kmeansj}) can be deployed to find groups. However, they do not satisfy the above condition of generalized superpixels (i.e., spatial connectivity). To ensure the resulting partitioning of generalized superpixel satisfies both the two above conditions, we develop a generalized superpixel segmentation algorithm (GSSA) in \cref{alg1}. Specifically, we consider the importance of corresponding coordinates $\mathbf{X}\in\mathbb{R}^{c\times n}$ (i.e., $\{{\mathbf x_i}\in\mathbb{R}^{c}\}_{i=1}^n$) of the point data $\mathbf{O}\in\mathbb{R}^{s\times n}$ (i.e., $\{{\textbf o_i}\in\mathbb{R}^{s}\}_{i=1}^n$) to satisfy the spatial connectivity. Our example results are shown in \cref{superpixel}.



With GSSA, the generalized superpixels that encode rich semantic information are ready as basic units instead of pixels in INRs.

\subsection{The Proposed S-INR}
To fully exploit the semantic information within and across the generalized superpixels, we propose a novel superpixel-informed implicit neural representation with elaborately designed modules, i.e., exclusive attention-based MLPs and a shared dictionary matrix. Next, we first introduce the basic structure of INR and then elaborate on our proposed S-INR.

INR maps input coordinates to corresponding output values parameterized by an MLP, which is typically continuous and differentiable. Specifically, for a given data $\mathbf{O}\in\mathbb{R}^{s\times n}$ (i.e., $\{{\textbf o_i}\in\mathbb{R}^{s}\}_{i=1}^n$), INR uses the MLP with the sinusoidal activation function to map its coordinate $\mathbf{X}\in\mathbb{R}^{c\times n}$ (i.e., $\{{\textbf x_i}\in\mathbb{R}^{c}\}_{i=1}^n$) to corresponding values $\hat{\mathbf{O}}\in\mathbb{R}^{s\times n}$ (i.e., $\{\hat{\textbf o}_i\in\mathbb{R}^{s}\}_{i=1}^n$), serving as the recovery results. The INR, mapping ${\mathbb R}^{c}$ to ${\mathbb R}^{s}$, can be formulated as follows:
\begin{equation}\label{inr}
	\Phi_{\theta}(\mathbf{x})=\mathbf{W}_L(\sin(\mathbf{W}_{L-1}(\cdots(\sin(\mathbf{W}_0\mathbf{x}+\mathbf{b}_0))\cdots)+\mathbf{b}_{L-1}))+\mathbf{b}_L,
\end{equation}where $\theta \triangleq \{{\mathbf W}_l, \mathbf{b}_{l}\}_{l=0}^{L}$ denotes the learnable parameters, $\{\mathbf{W}_l\in{\mathbb R}^{c_{l+1}\times c_l}\}_{l=0}^L$ are weight matrices, and $\{\mathbf{b}_l\in{\mathbb R}^{c_{l+1}}\}_{l=0}^L$ denote bias vectors. Here, $c_{L+1}=s$ is the output dimension and $L$ denotes the number of layers in INR.


Different from using pixels as basic units in INR, we suggest utilizing generalized superpixels as the basic units of S-INR, which encode rich inherent semantic information of the data. The details of S-INR are presented below.


First, we propose a standard S-INR to exploit the semantic information within and across generalized superpixels. Specifically, we suggest utilizing the shared dictionary matrix ${\mathbf D} \in {\mathbb R}^{s\times r}$, which can interact with each generalized superpixel. With ${\mathbf D}$, the mathematical expression of the standard S-INR is as follows:\begin{equation}\label{S-INR1}
	\hat{\mathbf{o}}^{k}={\mathbf D}(\Phi_{\theta_{k}}(\mathbf{x}^{k})),\; k= 1,\;\dots,\; K,
\end{equation}where our proposed S-INR maps the corresponding coordinate of the $k$-th generalized superpixels $\mathbf{X}^{k}\in\mathbb{R}^{c\times n_k}$ (i.e., $\{\mathbf{x}_{i}^{k}\in\mathbb{R}^{c}\}_{i=1}^{n_k}$) to corresponding values $\hat{\mathbf{O}}^{k}\in\mathbb{R}^{s\times n_k}$ (i.e., $\{\hat{\mathbf{o}}_{i}^{k}\in\mathbb{R}^{s}\}_{i=1}^{n_k}$), which is then aggregated to obtain recovery results $\hat{\mathbf{O}}\in\mathbb{R}^{s\times n}$ (i.e., $\{\hat{\mathbf{o}}_i\in\mathbb{R}^{s}\}_{i=1}^n$), $n_1+\cdots+n_K=n$. Here, $\Phi_{\theta_k}:{\mathbb R}^{c}\to {\mathbb R}^{r}$ denotes the $k$-th INR representing the $k$-th generalized superpixel with learnable parameters $\theta_{k} \triangleq \{{\mathbf W}_l^{k}, \mathbf{b}_{l}^{k}\}_{l=0}^{L}$ and $K$ denotes the number of generalized superpixels. Note that in our work we set the dictionary matrix $\mathbf{D}$ to be a learnable parameter matrix, which allows for more flexible representation learning compared to using a predefined coding matrix in the dictionary learning field. 



Second, we develop an advanced S-INR that significantly enhances the expressiveness of S-INR in representing each generalized superpixel. Concretely, we propose an exclusive attention-based MLP $\Psi_{\theta_{k}}$, which is tailored to capture correlations across feature dimensions within each generalized superpixel. With $\Psi_{\theta_{k}}$, the formal formulation of S-INR is formulated as follows:
\begin{equation}\label{S-INR2}
	\hat{\mathbf{o}}^{k}={\mathbf D}(\Psi_{\theta_{k}}(\mathbf{x}^{k})),\; k= 1,\;\dots,\;K.
\end{equation}Specifically, for the proposed exclusive attention-based MLP $\Psi_{\theta_k}$, we add a self-attention function $\psi^{k}_{l}:{\mathbb R}^{c_{l+1}}\to{\mathbb R}^{c_{l+1} }$ \cite{se} after the $l$-th layer in the the $k$-th INR $\Phi_{\theta_k}$, so that the output of the original $l$-th layer is further processed by $\psi_l^{k}$. Thus, the parameters of the advanced S-INR can be denoted as $\{\mathbf{D}, \;\{\theta_{k}\}_{k=1}^{K}\}$, where $\theta_{k} \triangleq\{{\mathbf U}_l^{k},\;{\mathbf V}_l^{k},\;{\mathbf W}_l^{k}, \; \mathbf{b}_{l}^{k}\}_{l=0}^{L}$ denotes the learnable parameters. The mathematical expression of $\Psi_{\theta_k}$ is formulated as follows:\begin{equation}\label{selayer}
	\begin{split}
		\Psi_{\theta_{k}}(\mathbf{x}^{k})=&\psi_{L}^{k}(\mathbf{W}_{L}^{k}(\psi_{L-1}^{k}(\operatorname{sin}(\mathbf{W}_{L-1}^{k}(\cdots\psi_{0}^{k}(\operatorname{sin}(\mathbf{W}_{0}^{k}\mathbf{x}^{k}+\mathbf{b}_{0}^{k}))\cdots)\\&+\mathbf{b}_{L-1}^{k})))+\mathbf{b}_{L}^{k}), \\
	\end{split}
\end{equation}where $\psi_l^{k}(\mathbf{z}_{l+1}^{k})= \eta(\mathbf{U}_{l}^{k}(\delta(\mathbf{V}_{l}^{k}(\tau(\mathbf{z}_{l+1}^{k}))))\otimes\mathbf{z}_{l+1}^{k},\; l=0,\;\dots\;, L$. Here, if $l=0$, ${\mathbf z}^{k}_{1}=\sin(\mathbf{W}^{k}_0\mathbf{x}^{k}+\mathbf{b}^{k}_0)$, if $l=1,\dots,L-1$, $\mathbf{z}^{k}_{l+1}=\sin(\mathbf{W}^{k}_l\mathbf{z}^{k}_l+\mathbf{b}^{k}_l)$, and if $l=L$, $\mathbf{z}^{k}_{l+1}=\mathbf{W}^{k}_l\mathbf{z}^{k}_l+\mathbf{b}^{k}_l$. The symbol $\tau(\cdot)$ denotes channel-wise average pooling, $\delta(\cdot)$ denotes the ReLU activation function, $\eta(\cdot)$ denotes the sigmoid function, $\otimes$ denotes the channel-wise product, and $\{{\mathbf{U}^{k}_l}\in{\mathbb R}^{c_{l+1}\times c_{l+1}}\}_{l=0}^{L}$ and $\{{\mathbf{V}^{k}_l}\in{\mathbb R}^{c_{l+1}\times c_{l+1}}\}_{l=0}^{L}$ are learnable parameter matrices.



In summary, our proposed S-INR fully exploits the semantic information within and across generalized superpixels. The elaborately designed S-INR successfully as an effective tool exploits the inherent semantic information of the data under the INR framework.

\subsection{The Proposed Recovery Model}
To examine the representation capacity of our proposed S-INR, we propose an S-INR-based data recovery model tailored for multi-dimensional data recovery tasks. This model can be formulated as follows:
\begin{equation}
	\label{loss}
	\begin{split}
		\min_{\mathbf{D}, \;\{\theta_k\}_{k=1}^{K}}\; \mathcal{L}(\mathbf{O}, \; \hat{\mathbf{O}})=\sum_{k=1}^{K}\sum_{i=1}^{n_k}\mathcal{L}({\mathbf o}_i^{k},\;{\mathbf D}(\Psi_{\theta_k}(\mathbf{x}_i^{k}))),
	\end{split}
\end{equation}where $\Psi_{\theta_k}(\cdot)$ denotes the exclusive self-attention MLP to fully exploit semantic information within corresponding generalized superpixel, ${\mathbf D}$ denotes the shared dictionary matrix that interacts with each generalized superpixel to capture commonalities between them, the generalized superpixel $\mathbf{O}^{k}\in\mathbb{R}^{s\times n_k}$ (i.e., $\{{\mathbf{o}^{k}_i}\in\mathbb{R}^{s}\}_{i=1}^{n_k}$) is obtained from the observation ${\mathbf O}\in\mathbb{R}^{s\times n}$ (i.e., $\{{\mathbf{o}_i}\in\mathbb{R}^{s}\}_{i=1}^{n}$), ${\mathbf{x}}_i^{k}$ is corresponding coordinate of the ${\mathbf{o}}_i^{k}$, $K$ is the number of generalized superpixels, and $\mathcal{L}$ denotes the loss function. The aggregation of $\{\hat{\mathbf{O}}^{k}=\{{\mathbf D}(\Psi_{\theta_k}(\mathbf{x}_i^{k}))\}_{i=1}^{n_k}\}_{k=1}^K$ is the final recovery result $\hat{\mathbf O}$. Next, with our proposed S-INR-based data recovery model, we consider three specific data recovery tasks:

\begin{enumerate}
	\item\textbf{Data reconstruction} aims at approximating the ground truth. The loss function is $\sum_{k=1}^{K}\sum_{i=1}^{n_k}\|{\mathbf o_{i}}^{k}-{\mathbf D}(\Psi_{\theta_k}(\mathbf{x}_{i}^{k}))\|^2$, where ${\mathbf o}_{i}^{k}$ is the $i$-th point from $k$-th clean generalized superpixel.
	
	\item\textbf{Data completion} aims at recovering the underlying patterns and structures from incomplete observations. The loss function is $\sum_{k=1}^{K}\sum_{i=1}^{n_k}\|({\mathbf o_{i}}^{k}-{\mathbf D}(\Psi_{\theta_k}(\mathbf{x}_{i}^{k})))_\Omega\|^2$, where ${\mathbf o}_{i}^{k}$ is the $i$-th point from $k$-th incomplete generalized superpixel and $\Omega$ is the support of the observed data.
	
	\item\textbf{Data denoising} aims at recovering the underlying patterns and structures from noisy corrupted observation. We consider Gaussian noise with different standard deviations. The loss function is $\sum_{k=1}^{K}\sum_{i=1}^{n_k}\|{\mathbf o_{i}}^{k}-{\mathbf D}(\Psi_{\theta_k}(\mathbf{x}_{i}^{k}))\|^2$, where ${\mathbf o}_{i}^{k}$ is the $i$-th point from $k$-th noisy corrupted generalized superpixel.
\end{enumerate}

The proposed S-INR-based data recovery model is unsupervised which solely requires the observed data without training dataset. To solve the highly nonconvex and nonlinear problem \cref{loss}, we use prevalent gradient descent methods, i.e., the efficient adaptive moment estimation algorithm (Adam) \cite {adam}, to update the parameters $\mathbf{D}$ and $\{\theta_k\}_{k=1}^{K}$. Note that in the data completion, we utilize our proposed GSSA to find generalized superpixels from the incomplete observation after interpolation.

\section{Experiments}
In this section, we evaluate the representation capacity of our proposed S-INR on a range of multi-dimensional data recovery tasks. We first evaluate image data for tasks such as image reconstruction, image completion, and image denoising. Then, we extend the evaluation to more general point data arising from real-world applications, including weather data completion and 3D surface completion tasks. All experiments are conducted on a computer equipped with an Intel(R) UHD Graphics 630 CPU and an RTX 2080 Ti GPU.

\subsection{Experimental Settings}
To evaluate the performance of S-INR in different tasks, we use several evaluation metrics in experiments. For tasks involving image data, we evaluate the results using the peak signal-to-noise ratio (PSNR) \cite{psnr} and the structural similarity index (SSIM) \cite{ssim}. For tasks involving point data, we report the normalized root mean square error (NRMSE) and the R-Square for evaluations.


\subsubsection{Image Data.} To comprehensively evaluate the performance of our proposed S-INR on image data, we conduct experiments on three important image processing tasks: image reconstruction, image completion, and image denoising. These tasks allow us to assess different capabilities of S-INR, including reconstructing complex data, completing missing data, and recovering noisy corrupted data. (1) {Image Representation:} We utilize a RGB image $\it{Kodim}$ $(512 \times 768 \times 3)$ from Kodak dataset\footnote{\url{https://r0k.us/graphics/kodak/}} and a hyperspectral image (HSI) $\it{Pavia}$ $(256 \times 256 \times 32)$ from dataset\footnote{\url{https://www.ehu.eus/ccwintco/index.php/Hyperspectral_Remote_Sensing_Scenes}} as testing data. (2) {Image Completion:} We use a multispectral image (MSI) {\it Mor} $(256 \times 256 \times 31)$ in the ICVL dataset\footnote{\url{https://icvl.cs.bgu.ac.il/hyperspectral/}} as testing data. We adopt the random sampling rates of 0.025 and 0.05, labeled as Case1 and Case2, respectively. (3) {Image Denoising:} We adopt a MSI {\it Lehavim} $(256 \times 256 \times 31)$ also in the ICVL dataset as testing data. We consider Gaussian noise with standard deviations of 0.15 and 0.2, labeled as Case1 and Case2, respectively. We compare our method with state-of-the-art INR methods including Fourier\cite{tancik2020fourfeat} (i.e., ReLU+Pos.Enc), Gauss\cite{eccv_activation}, WIRE\cite{WIRE}, and SIREN\cite{sitzmann2020implicit}. The image reconstruction task is also compared with DIP\cite{dip}.


\subsubsection{Point Data.}	We further evaluate S-INR on processing point data tasks: weather data completion and 3D surface completion. These allow us to assess the ability of S-INR to model unstructured point data under real-world conditions. (1) {3D Surface Completion}: We consider the 3D surface completion task with a six-dimensional point set  $(x, y, z)-(R, G, B)$ formed by $n$ points, to estimate the color information of a given point cloud $\mathbf{O} \in \mathbb{R}^{3 \times n}$ from its corresponding coordinate $\mathbf{X} \in \mathbb{R}^{3 \times n}$. We use three scenes, i.e., {\it Scene1}, {\it Scene2}, and {\it Scene3}, acquired with spacetime stereo\cite{spacetime}, from the SHOT website\footnote{\url{http://www.vision.deis.unibo.it/research/80-shot}}. We adopt the random sampling rates of 0.025, 0.05, 0.075, and 0.1 as four cases, respectively. (2) {Weather Data Completion}: The weather data we utilize includes five values, namely precipitation (Prcp), soil water evaporation (SWE), vapor pressure (VP), and extreme temperatures (Tmax and Tmin). We utilize three datasets\footnote{\url{https://daac.ornl.gov/cgi-bin/dsviewer.pl?ds id=2130}} from the North America located at ($63^{\circ}$N, $157^{\circ}$W), ($61^{\circ}$N, $141^{\circ}$W), and ($62^{\circ}$N, $149^{\circ}$W), respectively. We adopt the random sampling rates of 0.1, 0.15, 0.2, and 0.25 as four cases, respectively. For the above two tasks, we report the average quantitative results for their corresponding four cases, respectively. We compare our method with standard regression methods including $K$-neighbors regressor (KNR)\cite{knr}, decision tree (DT)\cite{dt}, and random forest (RF)\cite{rf}. Additionally, we compare our method with the classical INR method, i.e., SIREN \cite{sitzmann2020implicit}.


\subsubsection{Hyperparameters Settings.} The configuration of hyperparameters is crucial for both our proposed S-INR method and other comparison methods. For WIRE, as in the original paper, we set the size of the hidden layer to 300. For other INR methods, we set the size of the hidden layer to 256. Regarding the number of hidden layers, we set it to 5 for all INR methods. To ensure a fair comparison, we apply other hyperparameters from the original papers when available. Otherwise, we select optimal values within the ranges stated in the papers. For DIP, we set its hyperparameters as in the original paper. To be fair, we reduce the size of the hidden layer to keep its parameter size similar to other INR methods.

For our S-INR, we set the size of the hidden layer to 35 and the number of hidden layers to 5. Other important hyperparameters include the number of superpixels $K$, the weight $\alpha$ of GSSA, the initialization parameter $\omega_0$, the learning rate, and the size of the dictionary matrix $r$. The number of superpixels $K$, depending on the scale of the input data, is chosen from $\{15, 25, 50\}$. The weight $\alpha$ is selected from the set $\{1, 5, 20\}$, with the specific choice dependent on the data type. The initialization parameter $\omega_0$, uses the same weight initialization scheme for $\omega_0$ as SIREN\cite{sitzmann2020implicit}, but the tunable value of $\omega_0$ is selected from $\{30, 150, 300\}$, which is influenced by $K$. The learning rate is searched within $\{1 \times 10^{-5}, 5 \times 10^{-5}, 1 \times 10^{-4}, 5 \times 10^{-4}, 1 \times 10^{-3}\}$. The size of the dictionary matrix $r$ is set to a multiple of $s$, where $s$ is the output dimension of the data, and the scale factor is set to 5 for RGB images, 4 for point data, and $\{2,3,4\}$ for HSIs.

	\subsection{Experimental Results}
	\subsubsection{Image Reconstruction Results.}
\begin{figure*}[htbp!]
		\tiny
		\setlength{\tabcolsep}{0.9pt}
		\begin{center}
			\begin{tabular}{ccc}
				\includegraphics [width=0.28\textwidth]{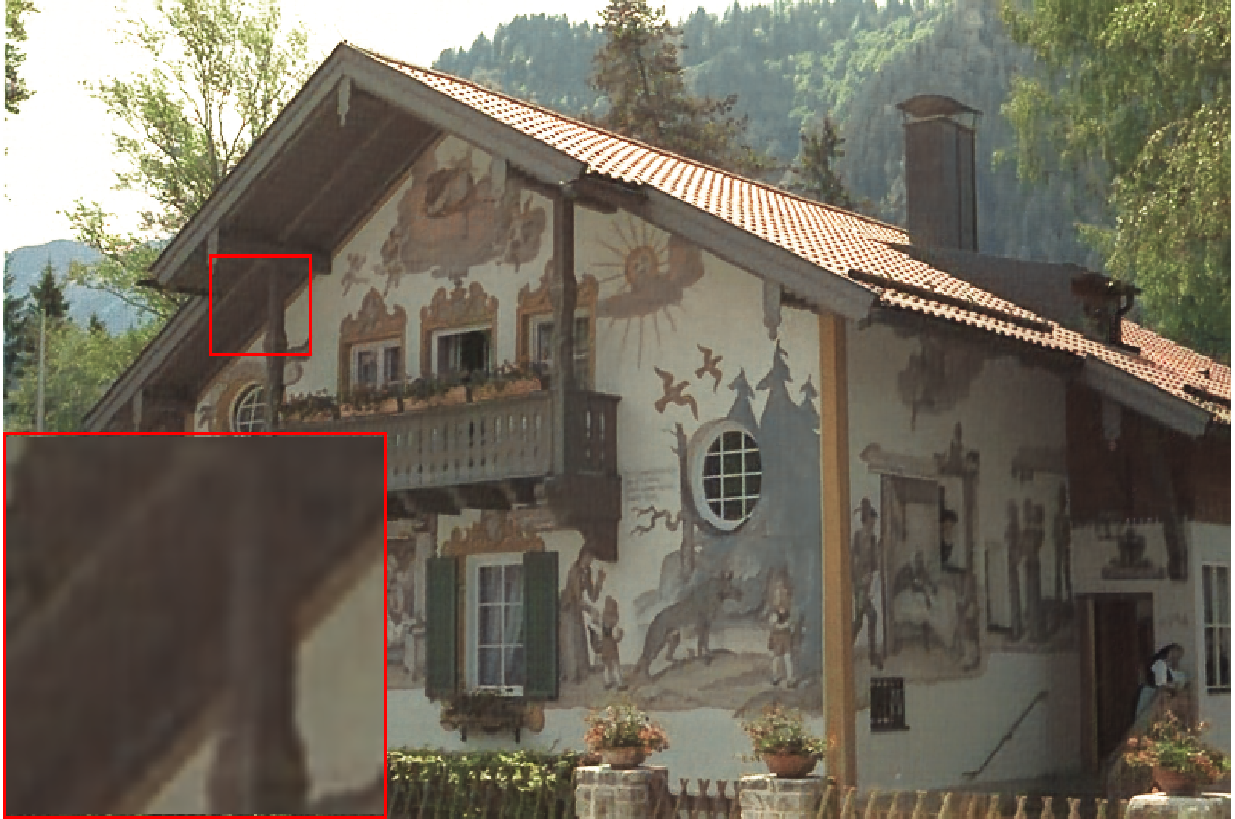}&
				\includegraphics [width=0.28\textwidth]{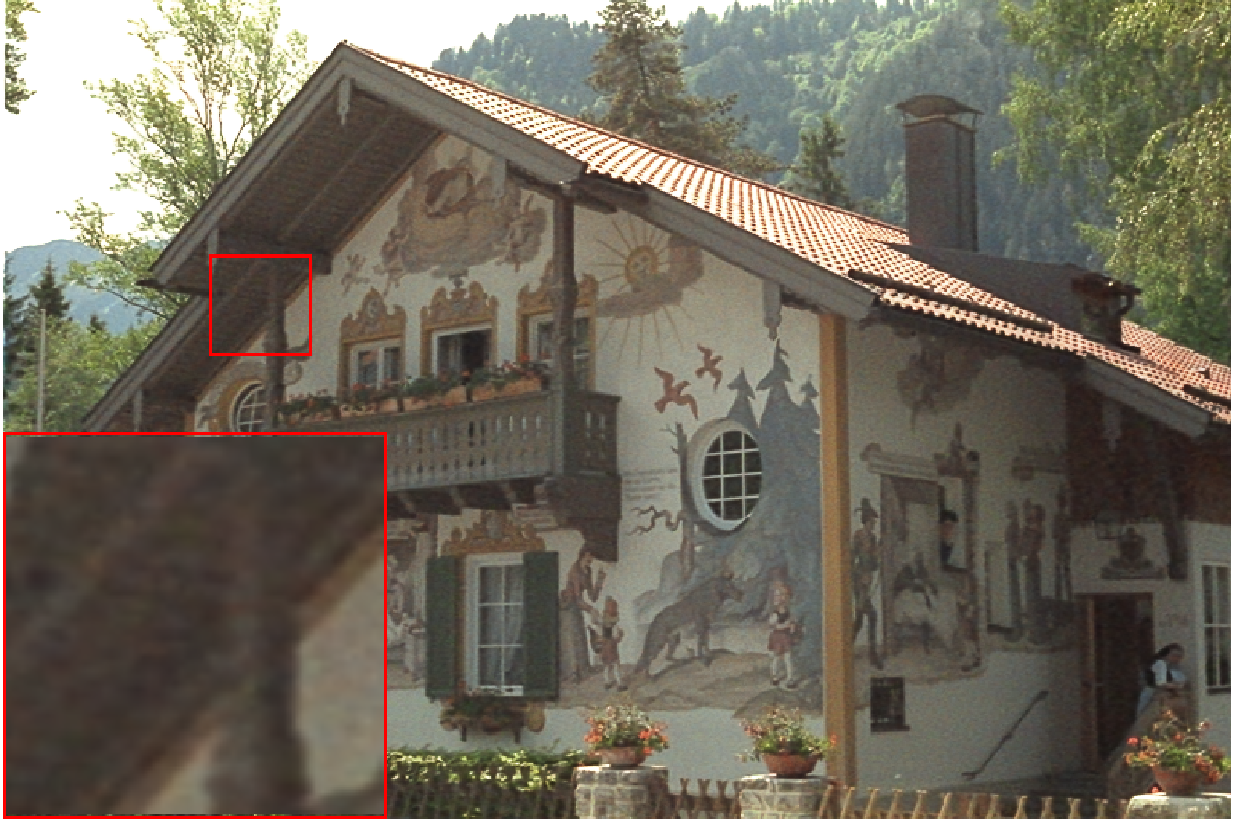}&
				\includegraphics [width=0.28\textwidth]{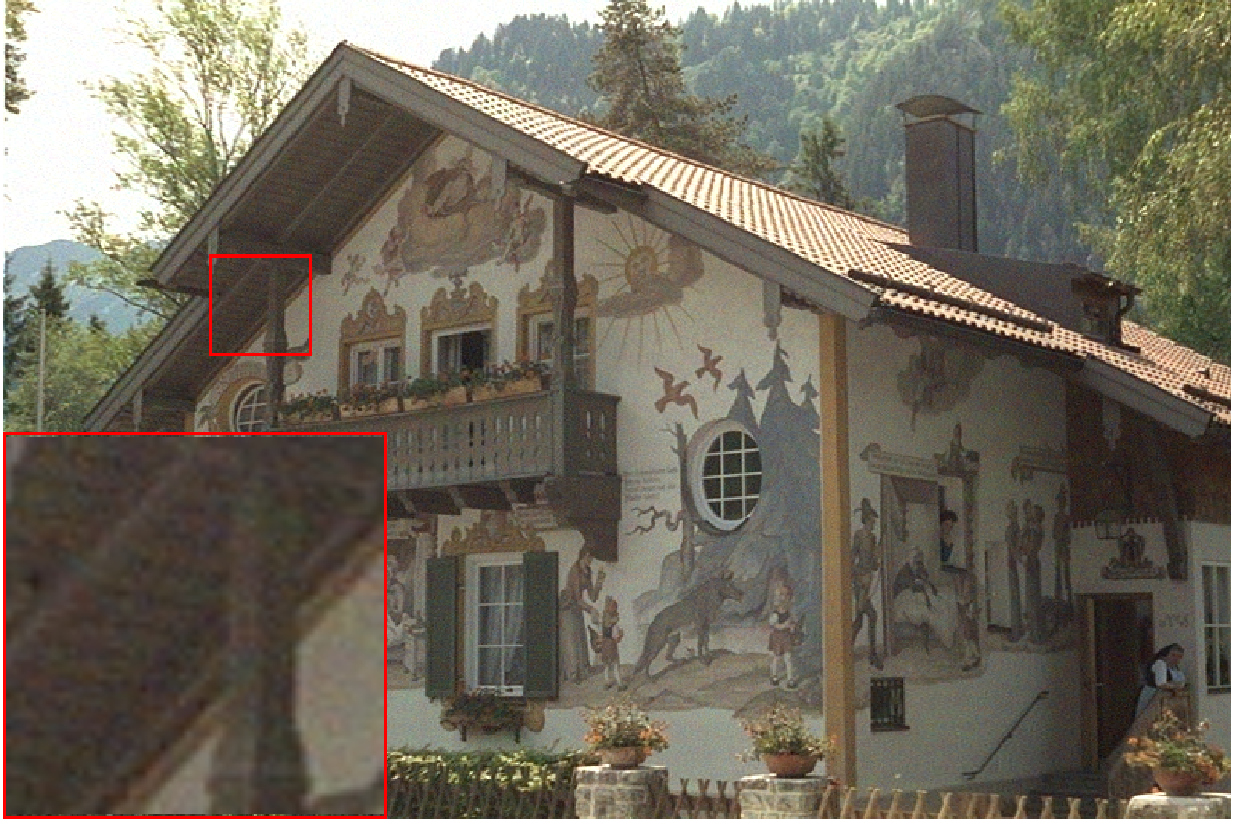}\\
				DIP & Fourier & Gauss \\
				\includegraphics [width=0.28\textwidth]{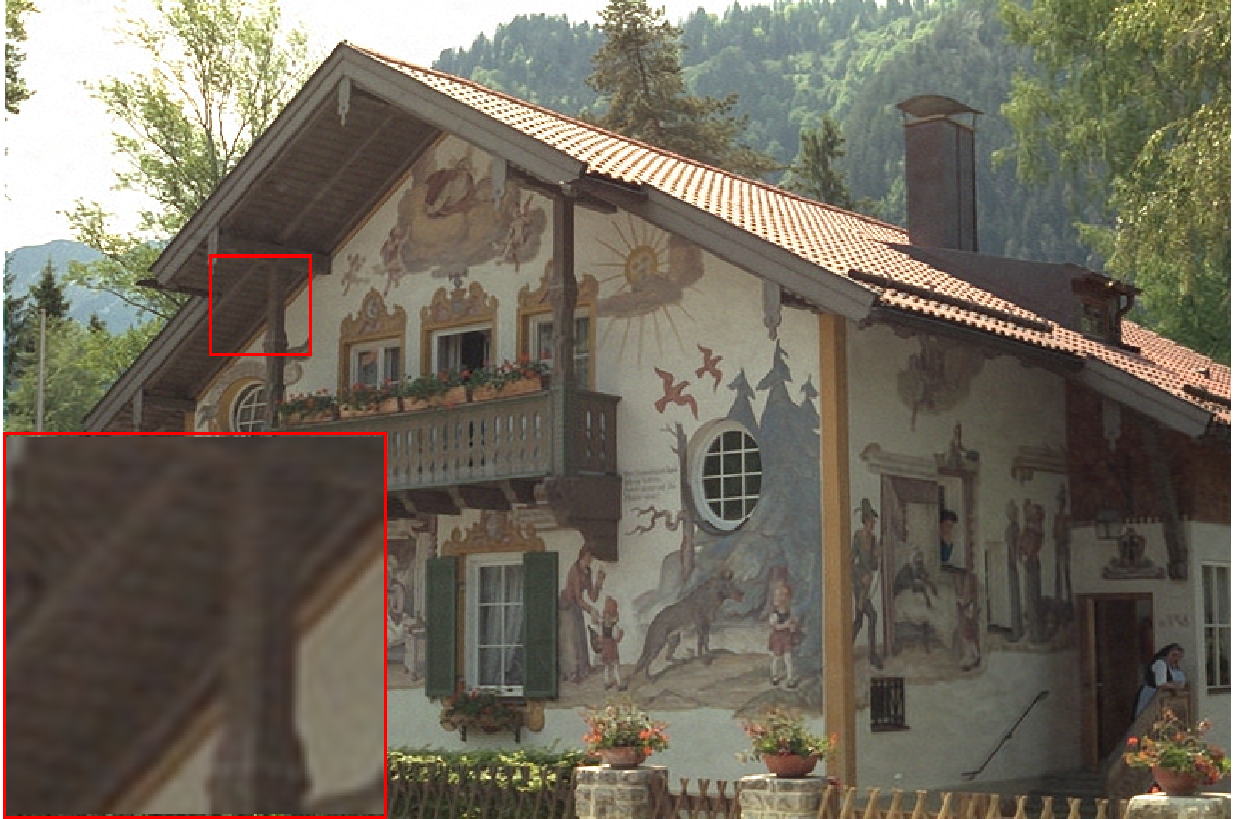}&
				\includegraphics [width=0.28\textwidth]{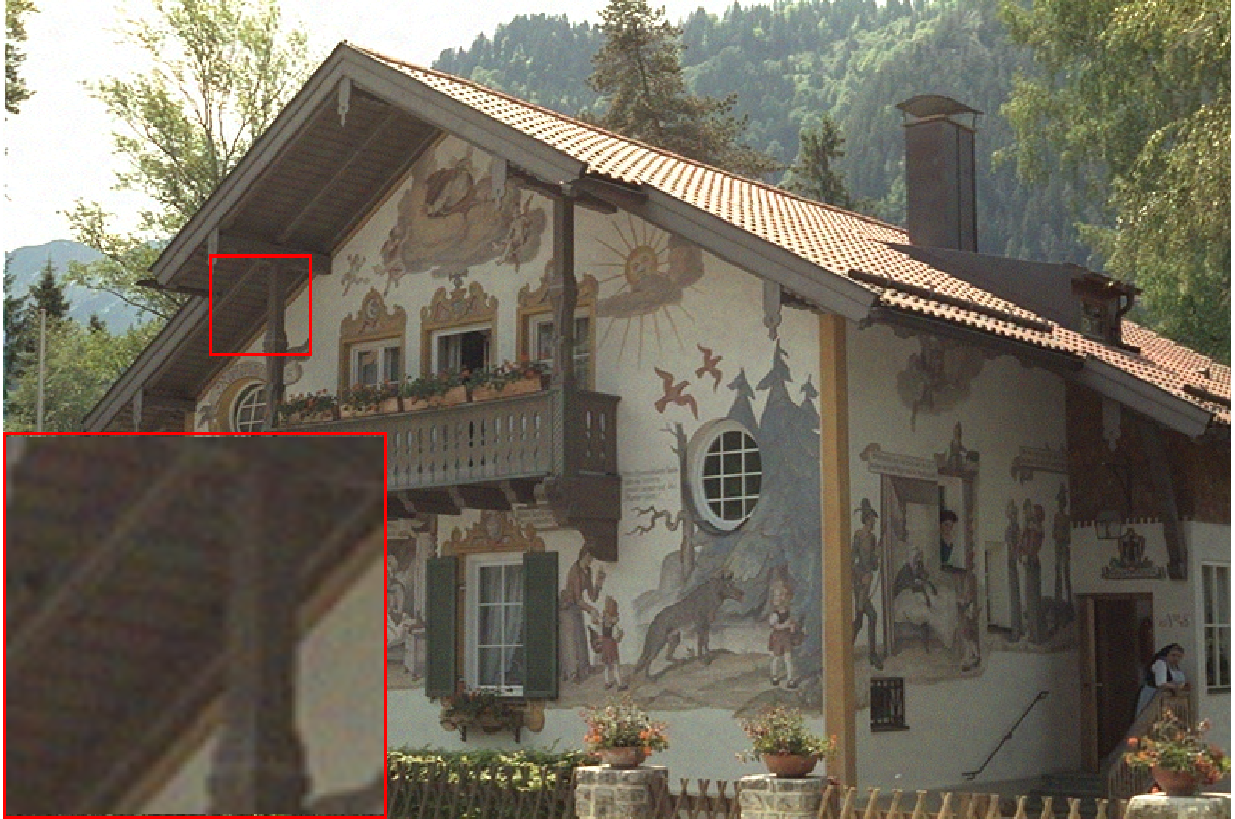}&
				\includegraphics [width=0.28\textwidth]{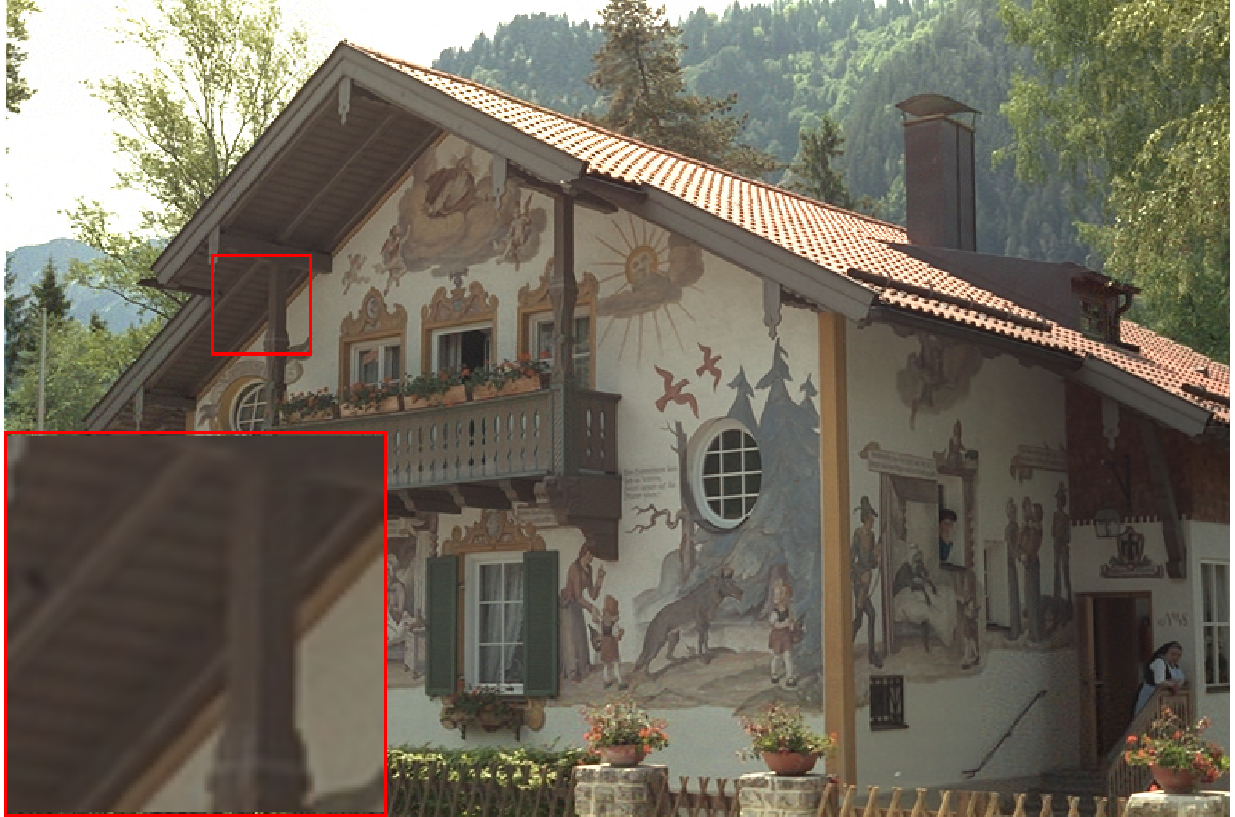}\\
				SIREN  & WIRE & S-INR \\ 
			\end{tabular}
		\end{center}
		\caption{The results of image reconstruction by different methods on RGB image $\it{Kodim}$.}
		\label{figfitting}
	\end{figure*}

The quantitative and qualitative results of image reconstruction are illustrated in   Table \ref{tablefitting} and Fig. \ref{figfitting}. We can observe that S-INR achieves superior PSNR and SSIM values compared to other comparison methods in Table \ref{tablefitting}. Meanwhile, we can observe that DIP, SIREN, and Fourier fail to represent image details, and the results of Gauss and WIRE are difficult to remove noise. In contrast, our method preserves rich textures and details without introducing noise, as shown in Fig. \ref{figfitting}. The promising reconstruction performance of our method can be largely attributed to the inherent semantic information captured by using the proposed generalized superpixels as basic units.
	
	\begin{table}[htbp!]
		\setlength{\tabcolsep}{0.9pt}
		\caption{The quantitative results for image reconstruction task. The \textbf{best} and \underline{second best} values are highlighted. (PSNR $\uparrow$, SSIM $\uparrow$)}
		\begin{center}
			\setlength{\tabcolsep}{6pt}			

			\resizebox{0.85\textwidth}{!}{\begin{tabular}{cccccccccccccc}
					\toprule
					Data& Metrics & \multicolumn{2}{c}{DIP} & \multicolumn{2}{c}{Fourier} & \multicolumn{2}{c}{Gauss} & \multicolumn{2}{c}{SIREN} & \multicolumn{2}{c}{WIRE} & \multicolumn{2}{c}{S-INR} \\ \midrule
					
					\multirow{2}{*}{\it{Kodim}}
					&PSNR & \multicolumn{2}{c}{30.154} & \multicolumn{2}{c}{32.101} & \multicolumn{2}{c}{30.188} & \multicolumn{2}{c}{33.052} & \multicolumn{2}{c}{$\underline{33.199}$} & \multicolumn{2}{c}{\textbf{36.077}}\\
					&SSIM & \multicolumn{2}{c}{0.882} & \multicolumn{2}{c}{0.899} & \multicolumn{2}{c}{0.862} & \multicolumn{2}{c}{$\underline{0.932}$} & \multicolumn{2}{c}{0.918} & \multicolumn{2}{c}{\textbf{0.965}} \\ \midrule
					\multirow{2}{*}{\it{Pavia}}
					&PSNR & \multicolumn{2}{c}{36.283} & \multicolumn{2}{c}{37.982} & \multicolumn{2}{c}{36.413} & \multicolumn{2}{c}{37.727} & \multicolumn{2}{c}{$\underline{38.455}$} & \multicolumn{2}{c}{\textbf{39.102}} \\
					&SSIM & \multicolumn{2}{c}{0.919} & \multicolumn{2}{c}{0.935} & \multicolumn{2}{c}{0.923} & \multicolumn{2}{c}{0.937} & \multicolumn{2}{c}{$\underline{0.941}$} & \multicolumn{2}{c}{\textbf{0.949}} \\
					
					\bottomrule
			\end{tabular}}
		\end{center}
		\label{tablefitting}

	\end{table}

	\subsubsection{Image Completion Results.}
	The quantitative and qualitative results of image completion are shown in Table \ref{tableinpainting} and the first row of Fig. \ref{figinde}. We can observe that our method obtains better quantitative results as compared with other INR methods in terms of PSNR and SSIM. From the visual results in Fig. \ref{figinde}, we can see that Fourier and SIREN fail to represent image details and produce blurry results, Gauss and WIRE show artifacts in the form of green discoloration and extraneous patterns. Meanwhile, S-INR can capture fine details without artifacts, validating the effectiveness of our method in image completion tasks. The superior performance of S-INR is attributed to its elaborately designed components to jointly exploit the inherent semantic information of the data.

	\begin{table}[htbp!]
		\caption{The quantitative results for image completion task. The \textbf{best} and \underline{second best} values are highlighted. (PSNR $\uparrow$, SSIM $\uparrow$)}
		\setlength{\tabcolsep}{0.9pt}
		\begin{center}
			\setlength{\tabcolsep}{3.5pt}

			\resizebox{0.85\textwidth}{!}{\begin{tabular}{ccccccccccccccc}
					\toprule
					Data&Cases & Metrics & \multicolumn{2}{c}{Observed} & \multicolumn{2}{c}{Fourier} & \multicolumn{2}{c}{Gauss} & \multicolumn{2}{c}{SIREN} & \multicolumn{2}{c}{WIRE} & \multicolumn{2}{c}{S-INR} \\ \midrule
					
					\multirow{4}{*}{\it{Mor}}&\multirow{2}{*}{Case1}
					&PSNR & \multicolumn{2}{c}{9.056} & \multicolumn{2}{c}{27.835} & \multicolumn{2}{c}{28.064} & \multicolumn{2}{c}{27.918} & \multicolumn{2}{c}{$\underline{28.087}$} & \multicolumn{2}{c}{\textbf{29.068}} \\
					&&SSIM & \multicolumn{2}{c}{0.030} & \multicolumn{2}{c}{0.801} & \multicolumn{2}{c}{$\underline{0.808}$} & \multicolumn{2}{c}{0.763} & \multicolumn{2}{c}{0.804} & \multicolumn{2}{c}{\textbf{0.823}} \\ 
					&\multirow{2}{*}{Case2}
					&PSNR & \multicolumn{2}{c}{9.169} & \multicolumn{2}{c}{31.017} & \multicolumn{2}{c}{31.026} & \multicolumn{2}{c}{30.929} & \multicolumn{2}{c}{$\underline{31.609}$} & \multicolumn{2}{c}{\textbf{32.281}} \\
					&&SSIM & \multicolumn{2}{c}{0.047} & \multicolumn{2}{c}{0.886} & \multicolumn{2}{c}{0.895} & \multicolumn{2}{c}{0.860} & \multicolumn{2}{c}{$\underline{0.897}$} & \multicolumn{2}{c}{\textbf{0.900}} \\
					\bottomrule
			\end{tabular}}
		\end{center}
		\label{tableinpainting}

	\end{table}	
	\begin{figure*}[htbp!]

		\tiny
		\setlength{\tabcolsep}{0.9pt}
		\begin{center}
			\begin{tabular}{ccccccc}
				\includegraphics [width=0.12\textwidth]{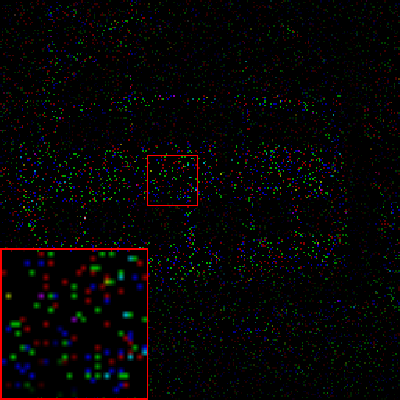}&
				\includegraphics [width=0.12\textwidth]{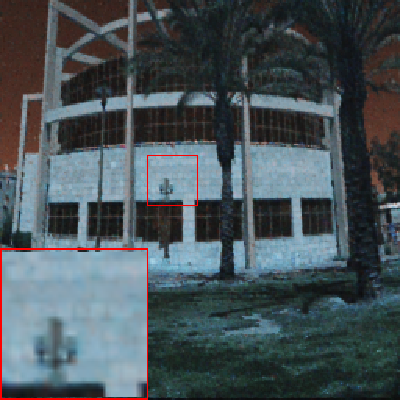}&	
				\includegraphics [width=0.12\textwidth]{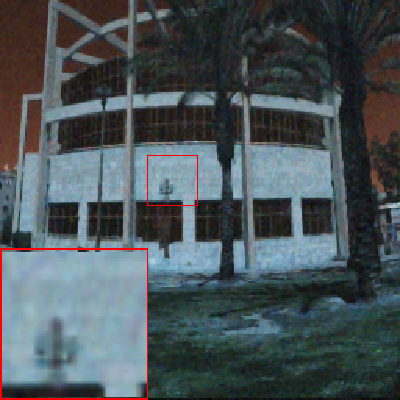}&
				\includegraphics [width=0.12\textwidth]{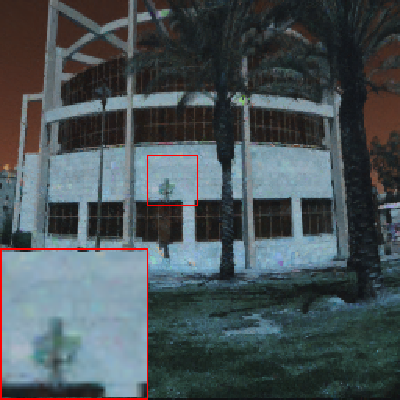}&
				\includegraphics [width=0.12\textwidth]{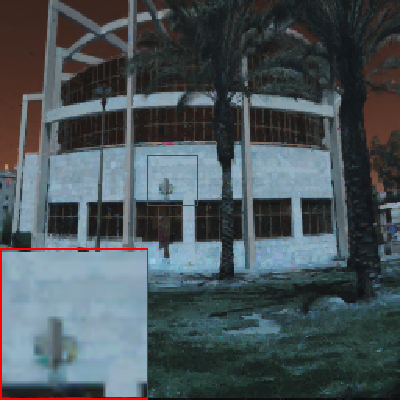}&
				\includegraphics [width=0.12\textwidth]{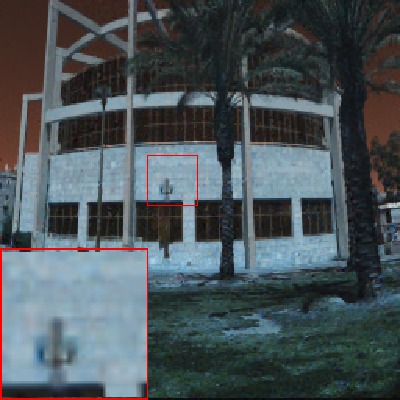}&
				\includegraphics [width=0.12\textwidth]{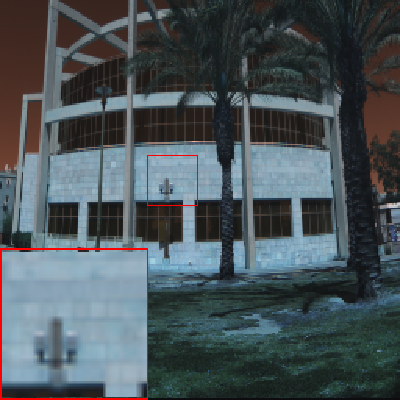}\\
				\includegraphics [width=0.12\textwidth]{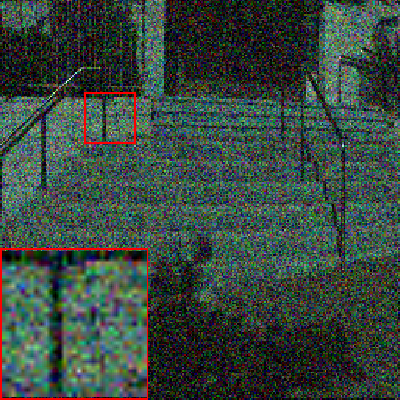}&
				\includegraphics [width=0.12\textwidth]{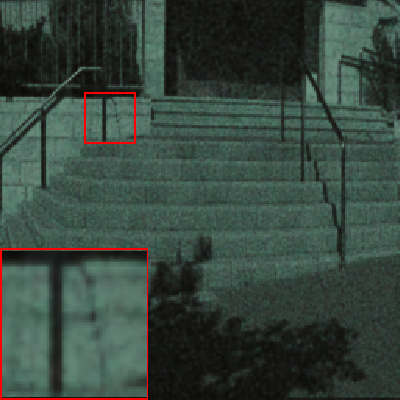}&
				\includegraphics [width=0.12\textwidth]{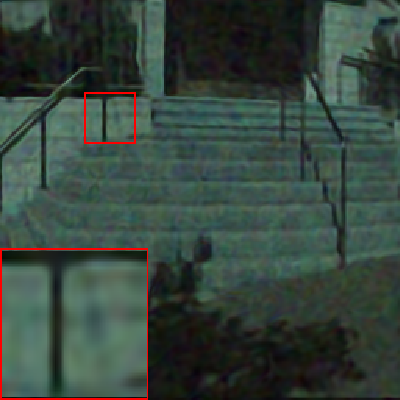}&			
				\includegraphics [width=0.12\textwidth]{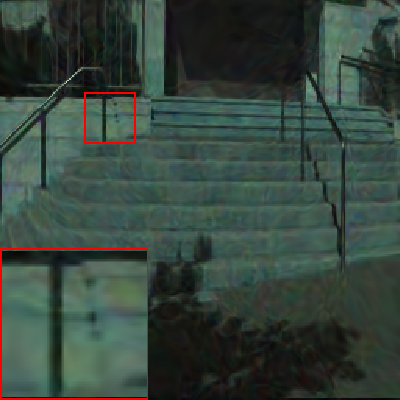}&
				\includegraphics [width=0.12\textwidth]{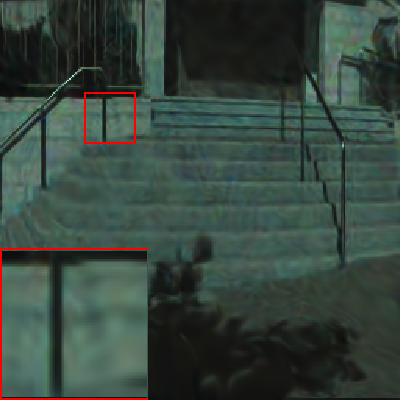}&
				\includegraphics [width=0.12\textwidth]{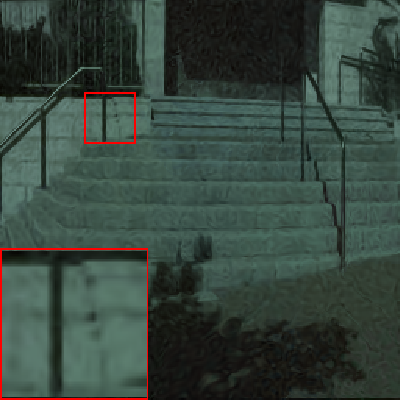}&
				\includegraphics [width=0.12\textwidth]{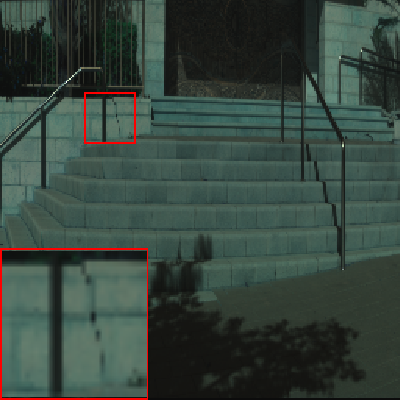}\\

				Observed & Fourier & SIREN & Gauss & WIRE & S-INR & Original\\
			\end{tabular}
		\end{center}

		\caption{From top to bottom list the results of image completion and image denoising by different methods on $\it{Mor}$ in Case2 and $\it{Lehavim}$ in Case1, respectively.}
		\label{figinde}

	\end{figure*}
    
    	\begin{table}[htbp!]
    	\setlength{\tabcolsep}{0.9pt}
    	\caption{The quantitative results for image denoising task. The \textbf{best} and \underline{second best} values are highlighted. (PSNR $\uparrow$, SSIM $\uparrow$)}
    	\begin{center}
    		\setlength{\tabcolsep}{3.5pt}
    		
    		\resizebox{0.85\textwidth}{!}{\begin{tabular}{ccccccccccccccc}
    				\toprule
    				Data&Cases & Metrics & \multicolumn{2}{c}{Observed} & \multicolumn{2}{c}{Fourier} & \multicolumn{2}{c}{Gauss} & \multicolumn{2}{c}{SIREN} & \multicolumn{2}{c}{WIRE} & \multicolumn{2}{c}{S-INR} \\ \midrule
    				
    				\multirow{4}{*}{\it{Lehavim}}&\multirow{2}{*}{Case1}
    				&PSNR & \multicolumn{2}{c}{16.479} & \multicolumn{2}{c}{32.514} & \multicolumn{2}{c}{32.853} & \multicolumn{2}{c}{30.536} & \multicolumn{2}{c}{$\underline{33.196}$} & \multicolumn{2}{c}{\textbf{33.696}} \\
    				&&SSIM & \multicolumn{2}{c}{0.224} & \multicolumn{2}{c}{0.832} & \multicolumn{2}{c}{0.856} & \multicolumn{2}{c}{0.784} & \multicolumn{2}{c}{$\underline{0.863}$} & \multicolumn{2}{c}{\textbf{0.867}} \\ 
    				&\multirow{2}{*}{Case2}
    				&PSNR & \multicolumn{2}{c}{13.977} & \multicolumn{2}{c}{$\underline{31.333}$} & \multicolumn{2}{c}{31.303} & \multicolumn{2}{c}{29.348} & \multicolumn{2}{c}{30.123} & \multicolumn{2}{c}{\textbf{32.394}} \\
    				&&SSIM & \multicolumn{2}{c}{0.144} & \multicolumn{2}{c}{0.796} & \multicolumn{2}{c}{$\underline{0.819}$} & \multicolumn{2}{c}{0.745} & \multicolumn{2}{c}{0.792} & \multicolumn{2}{c}{\textbf{0.831}} \\
    				\bottomrule
    		\end{tabular}}
    	\end{center}
    	\label{tabledenoising}
    	
    \end{table}		
    
	\subsubsection{Image Denoising Results.}
	The qualitative and quantitative results of image denoising are shown in  the second row of Fig. \ref{figinde} and Table \ref{tabledenoising}. The visual quality of our method is apparent, particularly in its ability to represent intricate details and remove heavy noise, such as the split on the wall reflected in the zoomed views, as shown in Fig. \ref{figinde}. On the contrary, Gauss, SIREN, and WIRE fail to remove all the noise, while Fourier has chromatic aberration and introduces background noise. In addition, as compared with other INR methods, our method achieves competitive PSNR and SSIM values in Table \ref{tabledenoising}. This indicates that S-INR is not only superior in the accuracy of visual quality but also in the quantitative metrics compared to other comparison methods. We attribute the ability to recover noisy images even without any regularization to the expressiveness and robustness of S-INR. It is achieved by respecting the individuality of each generalized superpixel and capturing the commonalities between them.

	%
	%
	\begin{table}[htbp!]

		\caption{The average quantitative results by different methods for point data recovery tasks. The \textbf{best} and \underline{second best} values are highlighted. (NRMSE $\downarrow$, R-Square $\uparrow$)}
		\resizebox{\textwidth}{!}{
			\begin{tabular}{ccccccccccccc}
				\toprule
				\multirow{2}{*}{Data} & \multicolumn{6}{c}{Weather Data Completion} & \multicolumn{6}{c}{3D Surface Completion} \\ \cmidrule(lr){2-7} \cmidrule(l){8-13}
				& \multicolumn{2}{c}{($63^{\circ}$N, $157^{\circ}$W)} & \multicolumn{2}{c}{($61^{\circ}$N, $141^{\circ}$W)} & \multicolumn{2}{c}{($62^{\circ}$N, $149^{\circ}$W)} & \multicolumn{2}{c}{\it{Scene1}} & \multicolumn{2}{c}{\it{Scene2}} & \multicolumn{2}{c}{\it{Scene3}} 
				\\ \cmidrule(lr){1-13}
				Method & NRMSE & R-Square & NRMSE  & R-Square& NRMSE & R-Square & NRMSE & R-Square & NRMSE  & R-Square& NRMSE & R-Square   \\ \midrule
				DT     & 0.101 & 0.698  & 0.149  &0.484&0.131& 0.486 & 0.171 &0.703 &0.126  &0.761 &0.141   & 0.721\\ 
				KNR    & $\underline{0.072}$ & $\underline{0.849}$ & 0.099 &0.766& $\underline{0.093}$& 0.739 & 0.112 &0.868 &0.093 &0.867 &0.094  & 0.874\\ 
				RF     & 0.076 & 0.829 &  0.111 & 0.712 &0.099   & 0.705 & $\underline{0.107}$ &$\underline{0.880}$ &$\underline{0.090}$ &0.875 &0.092  & 0.879\\ 
				SIREN  & 0.078  & 0.818  &$\underline{0.096}$  & $\underline{0.776}$ & $\underline{0.093}$& $\underline{0.741}$ & 0.109 &0.878 &$\underline{0.090}$ &$\underline{0.886}$ &$\underline{0.087}$  & $\underline{0.885}$\\ 
				S-INR   & \textbf{0.058}& \textbf{0.900} &\textbf{0.070}&\textbf{0.877} &\textbf{0.062}&\textbf{ 0.889} &\textbf{0.074} &\textbf{0.944} &\textbf{0.063} & \textbf{0.945}&\textbf{0.060}&\textbf{0.945}\\ 
				\bottomrule
			\end{tabular}
		}
		\label{Multidatatable}

	\end{table}

	
	\begin{figure*}
		
		\centering
		\includegraphics[width=1.0\linewidth]{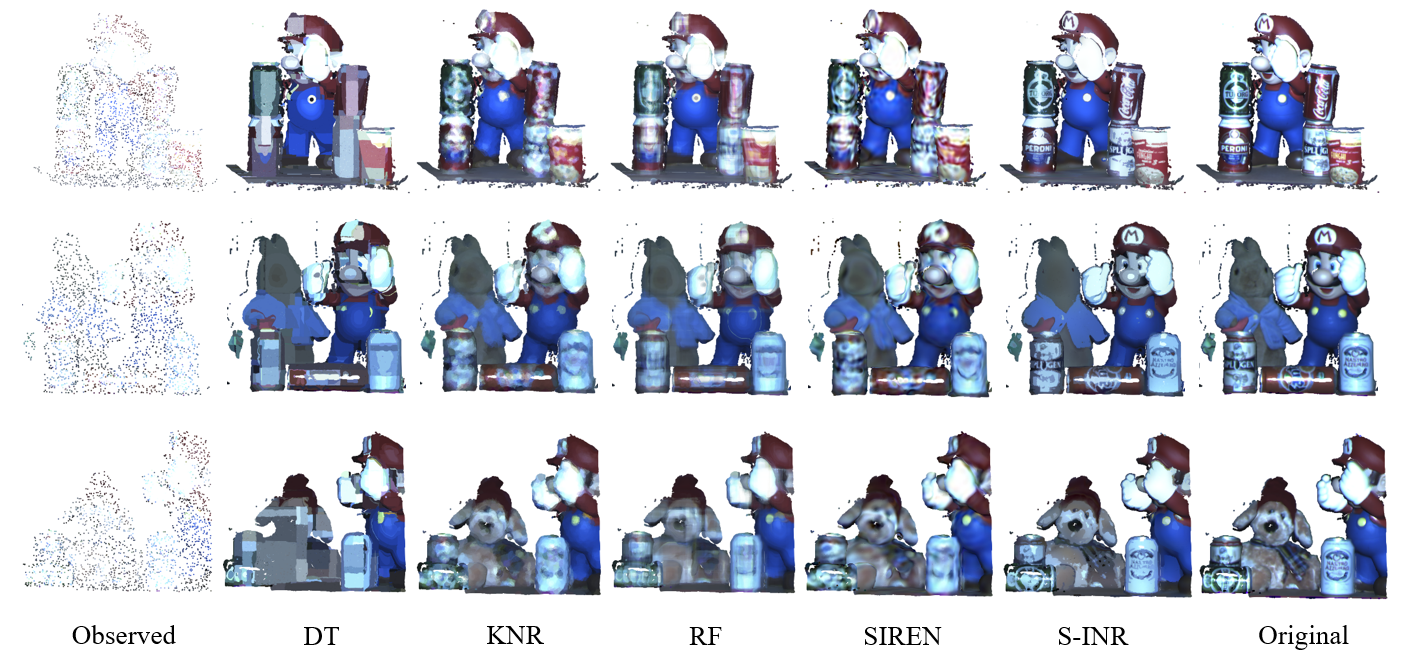}
		\caption{From top to down respectively list the results of 3D surface completion recovery by different methods for the 3D surface data $\it{Scene1}$, $\it{Scene2}$, and $\it{Scene3}$ with the random sampling rate of 0.025.}
		\label{fig:point3d}
		
	\end{figure*}
	
	\subsubsection{3D Surface Completion Results.}	
	The quantitative and qualitative results of the 3D surface completion are shown in Table \ref{Multidatatable} and Fig. \ref{fig:point3d}. We can observe that our S-INR achieves higher performance than other comparison methods in terms of NRMSE and R-Square. From the visual results, we can see that SIREN and other regression methods have difficulty in accurately recovering intricate details. In contrast, the proposed method can better represent the details due to its ability to exploit the inherent semantic information of the data, as evidenced by the patterns of the bottle in Fig. \ref{fig:point3d}. These results confirm the effectiveness and superiority of our S-INR for complex point data recovery tasks beyond images.

    	\begin{figure*}[htbp!]
    	\tiny
    	\setlength{\tabcolsep}{0.9pt}
    	\begin{center}
    		\begin{tabular}{ccccccc}
    			\includegraphics [width=0.12\textwidth]{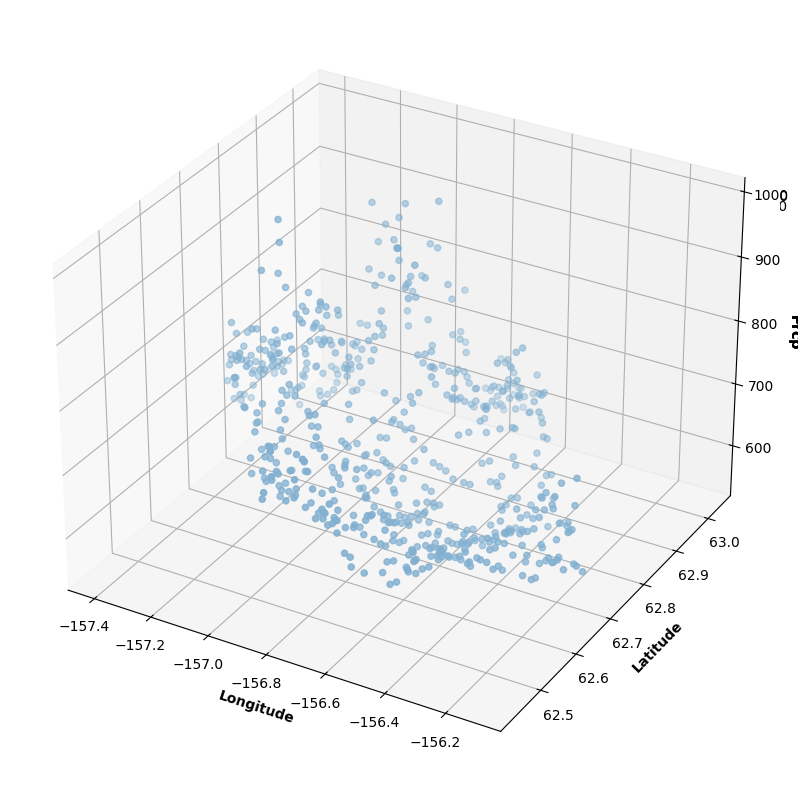}&
    			\includegraphics [width=0.12\textwidth]{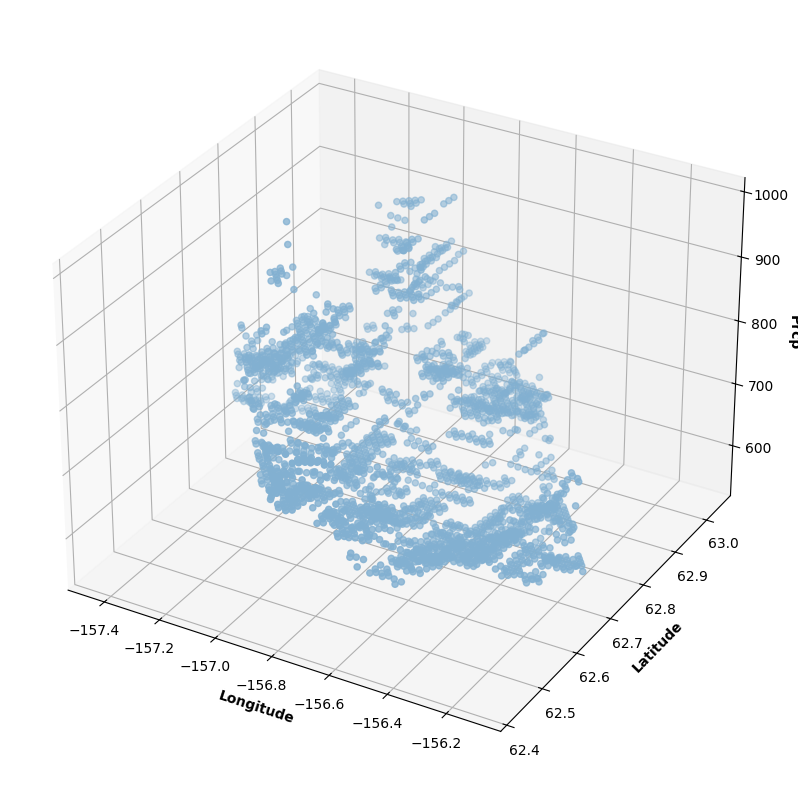}&
    			\includegraphics [width=0.12\textwidth]{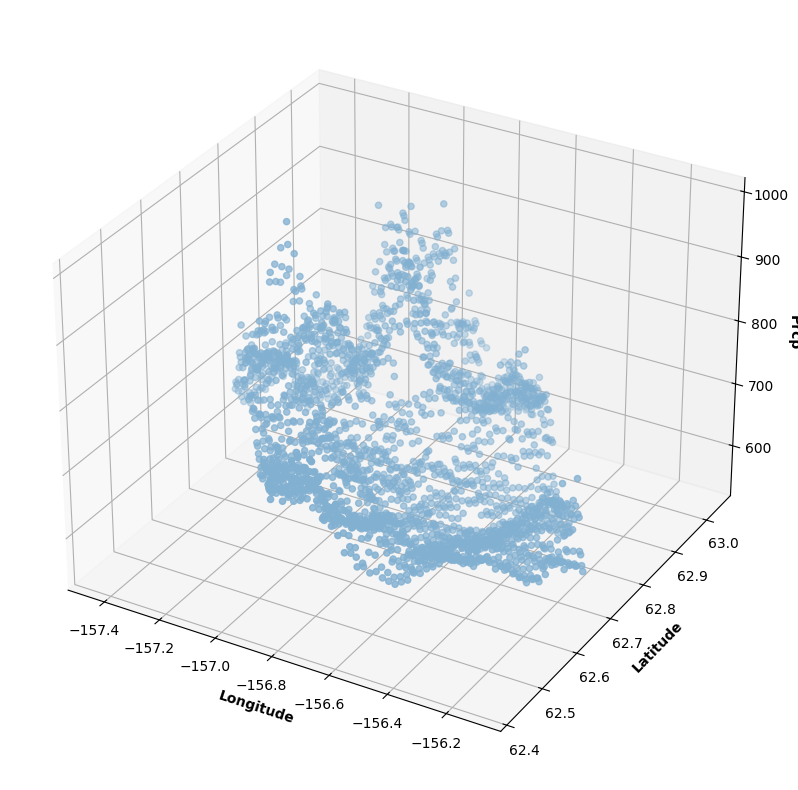}&
    			\includegraphics [width=0.12\textwidth]{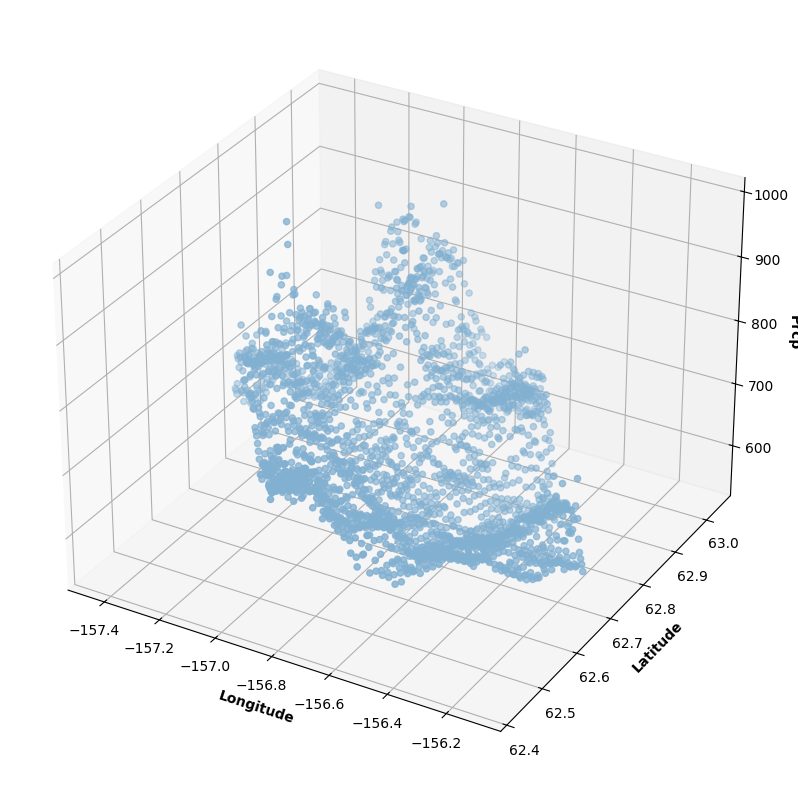}&
    			\includegraphics [width=0.12\textwidth]{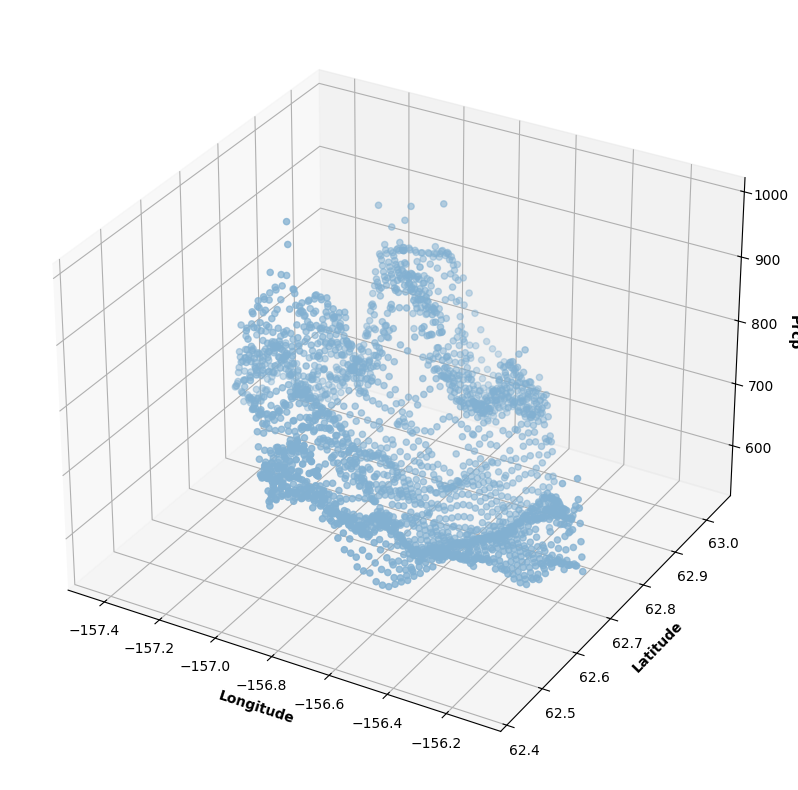}&
    			\includegraphics [width=0.12\textwidth]{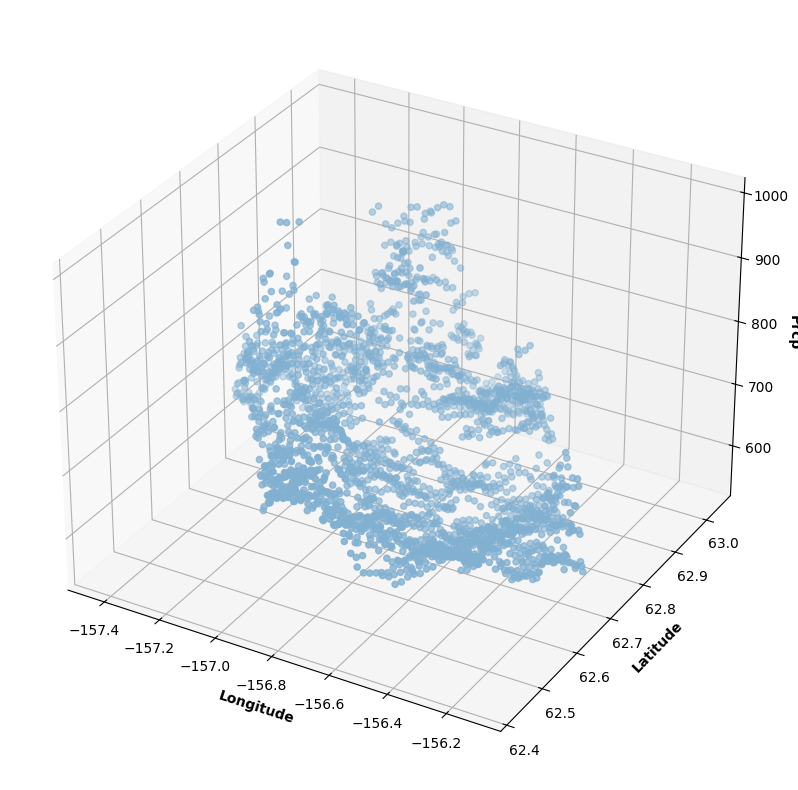}&
    			\includegraphics [width=0.12\textwidth]{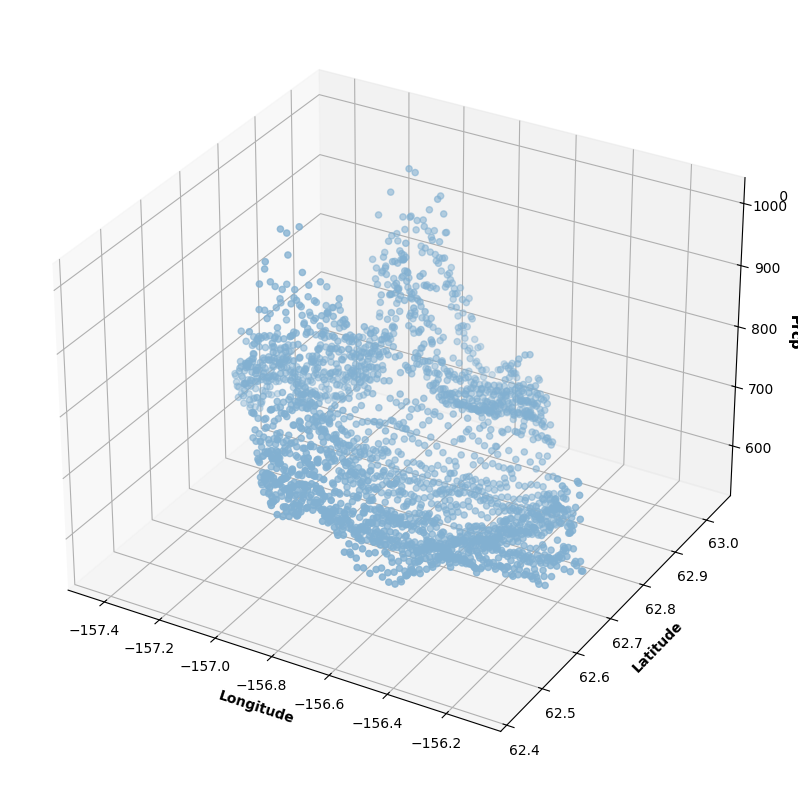}\\
    			\includegraphics [width=0.12\textwidth]{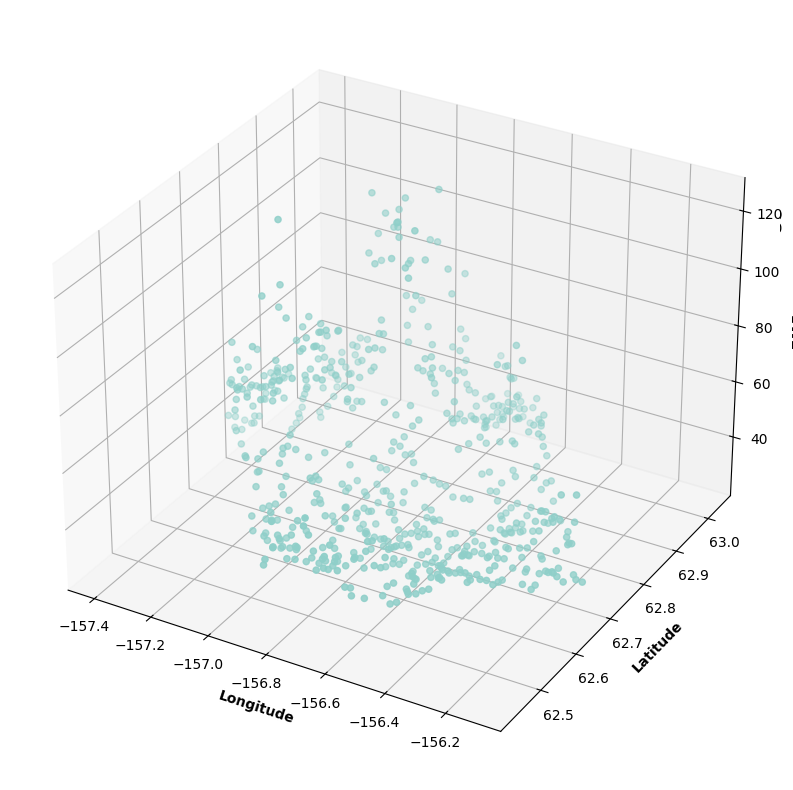}&
    			\includegraphics [width=0.12\textwidth]{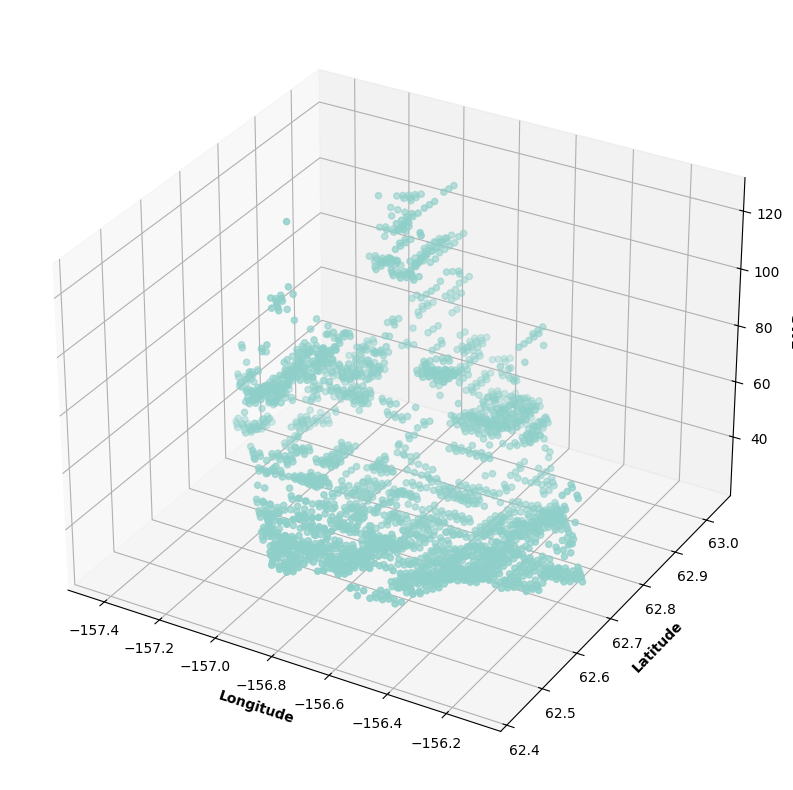}&
    			\includegraphics [width=0.12\textwidth]{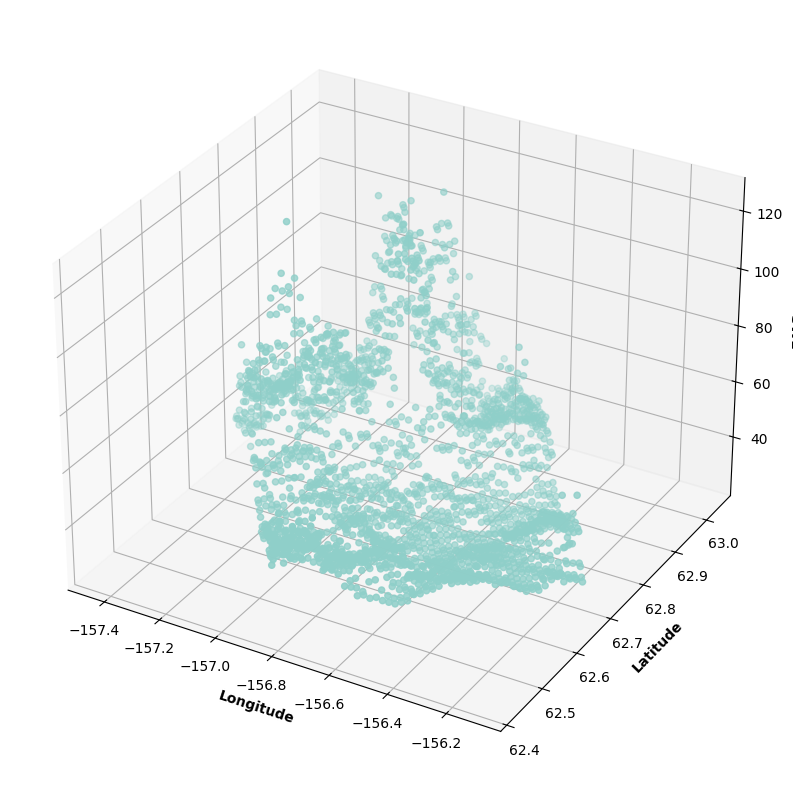}&
    			\includegraphics [width=0.12\textwidth]{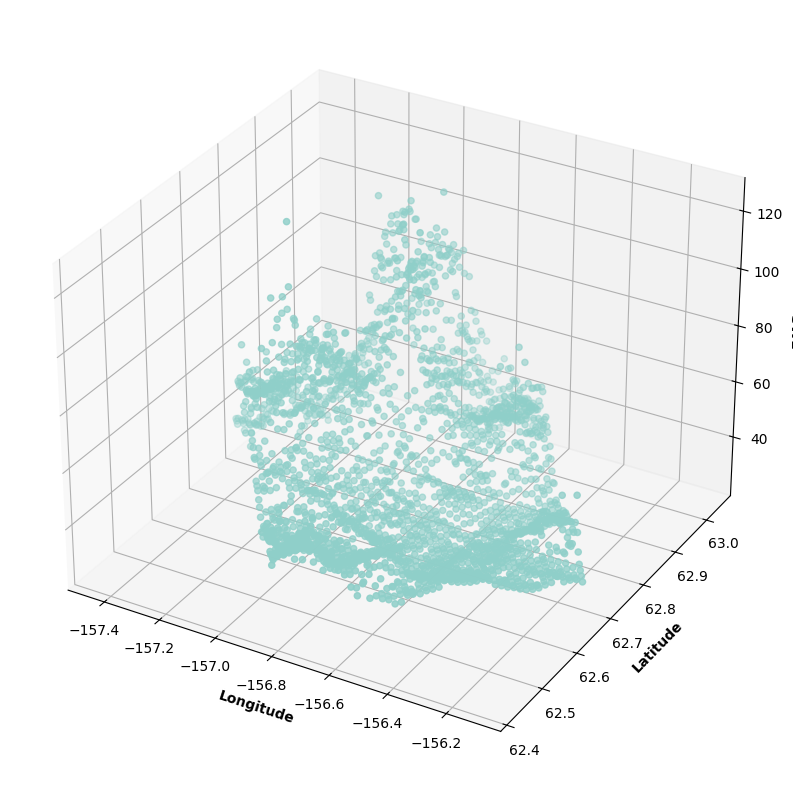}&
    			\includegraphics [width=0.12\textwidth]{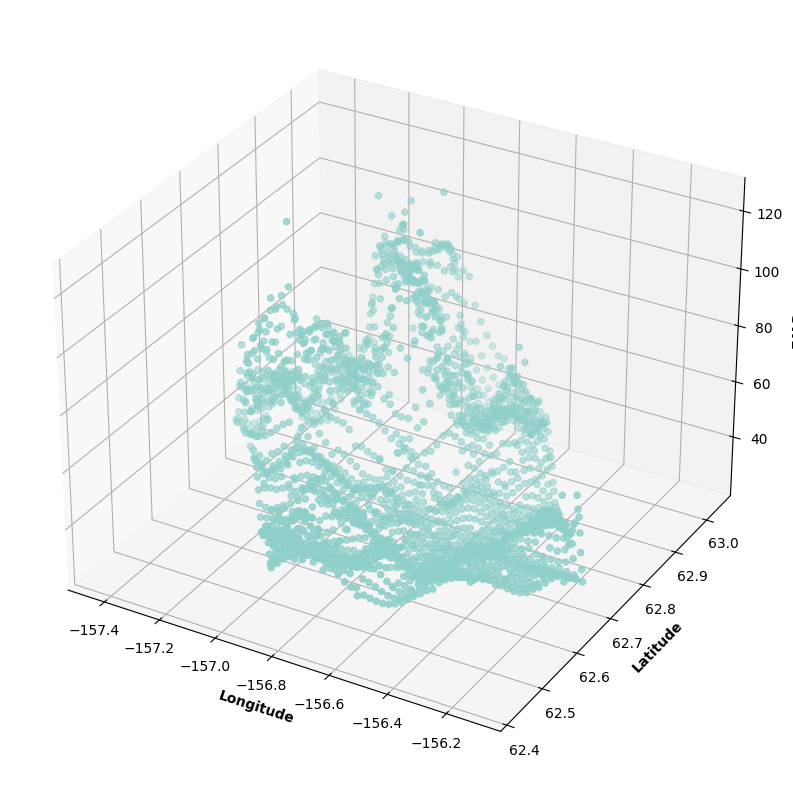}&
    			\includegraphics [width=0.12\textwidth]{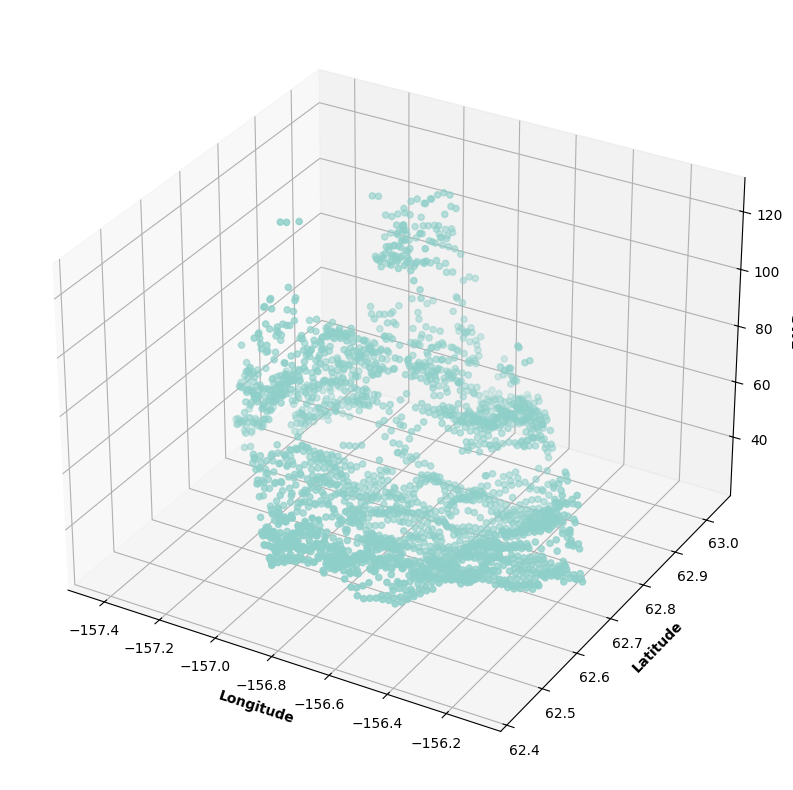}&
    			\includegraphics [width=0.12\textwidth]{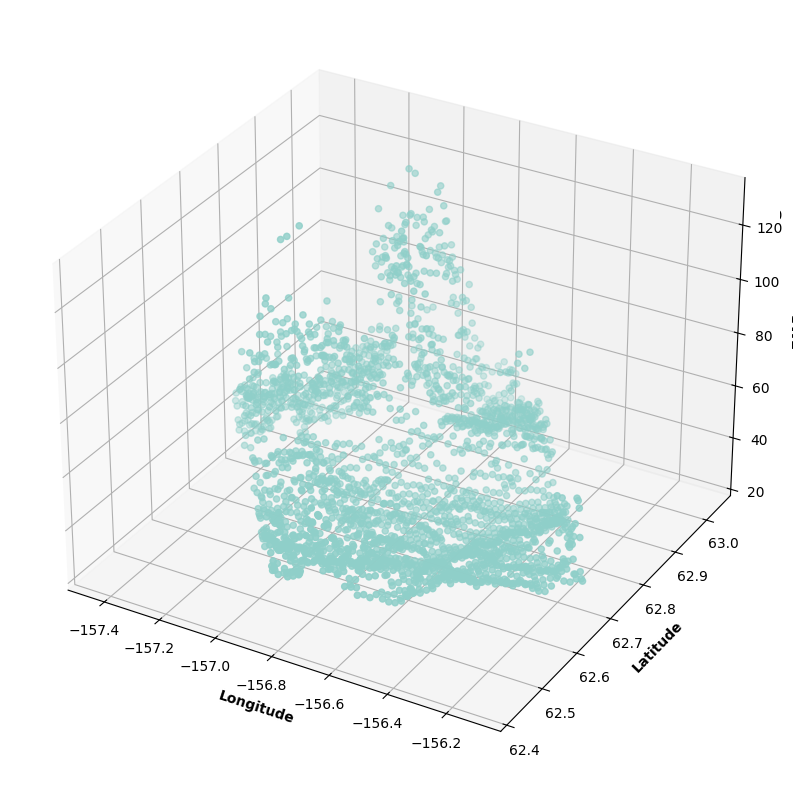}\\
    			\includegraphics [width=0.12\textwidth]{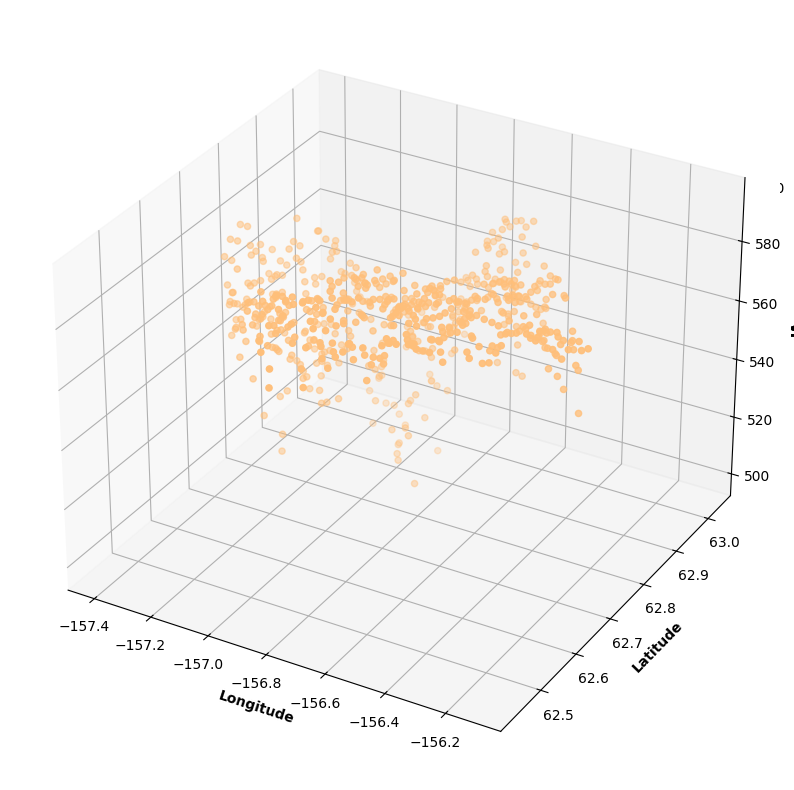}&
    			\includegraphics [width=0.12\textwidth]{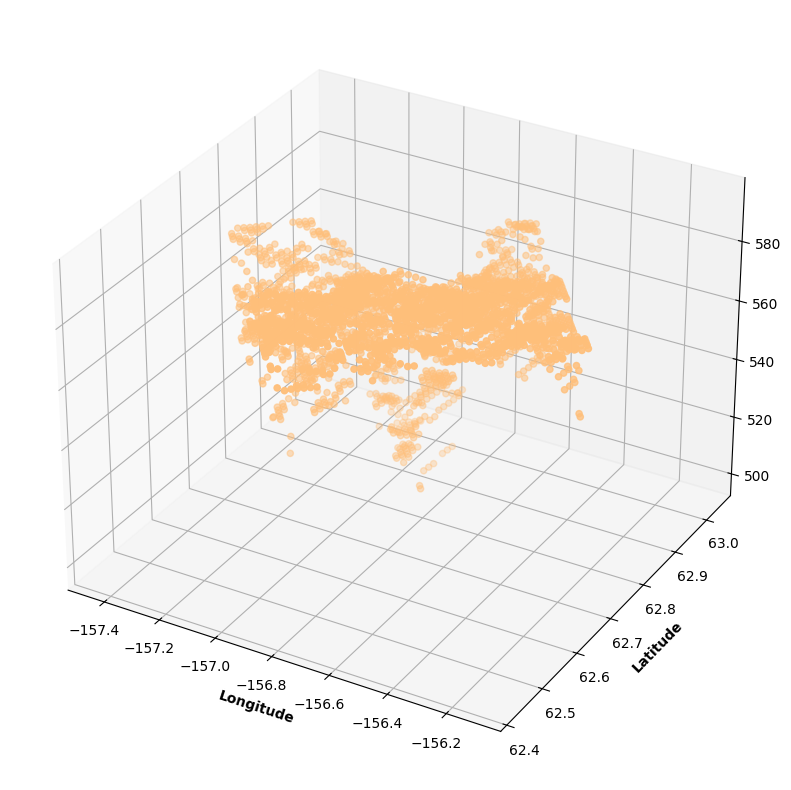}&
    			\includegraphics [width=0.12\textwidth]{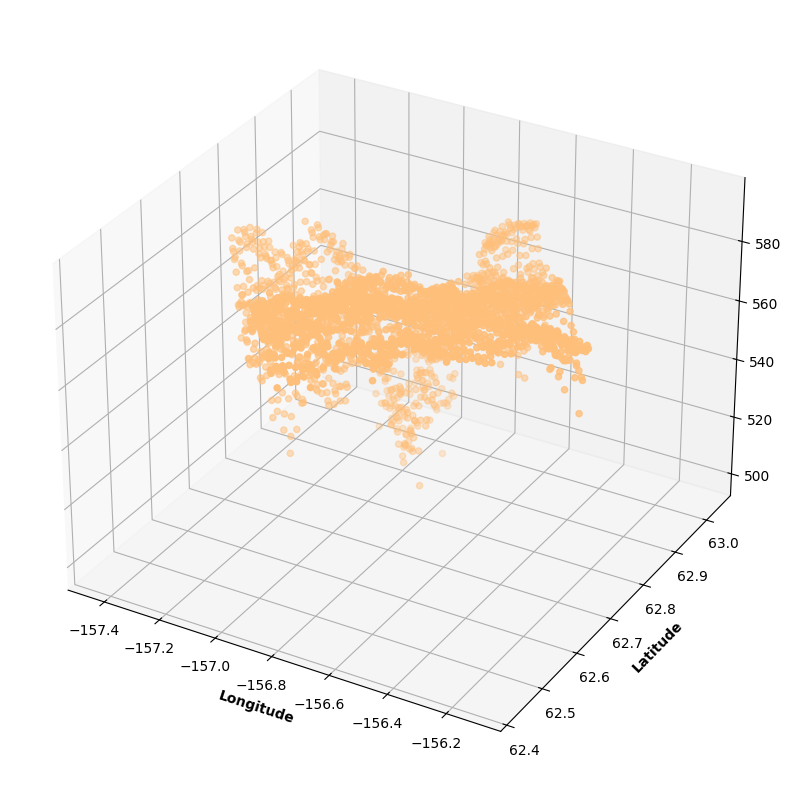}&
    			\includegraphics [width=0.12\textwidth]{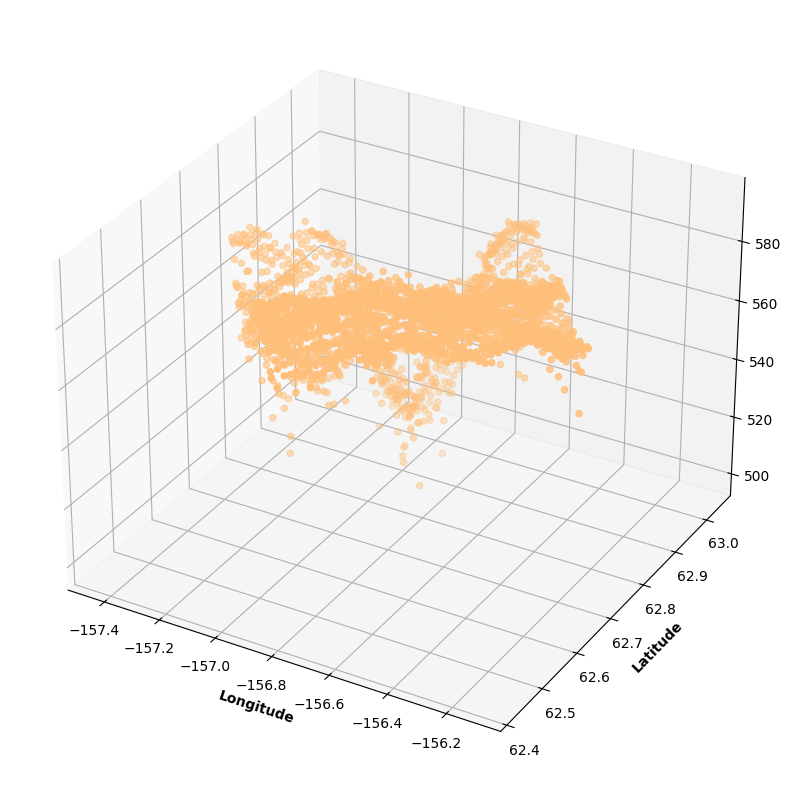}&
    			\includegraphics [width=0.12\textwidth]{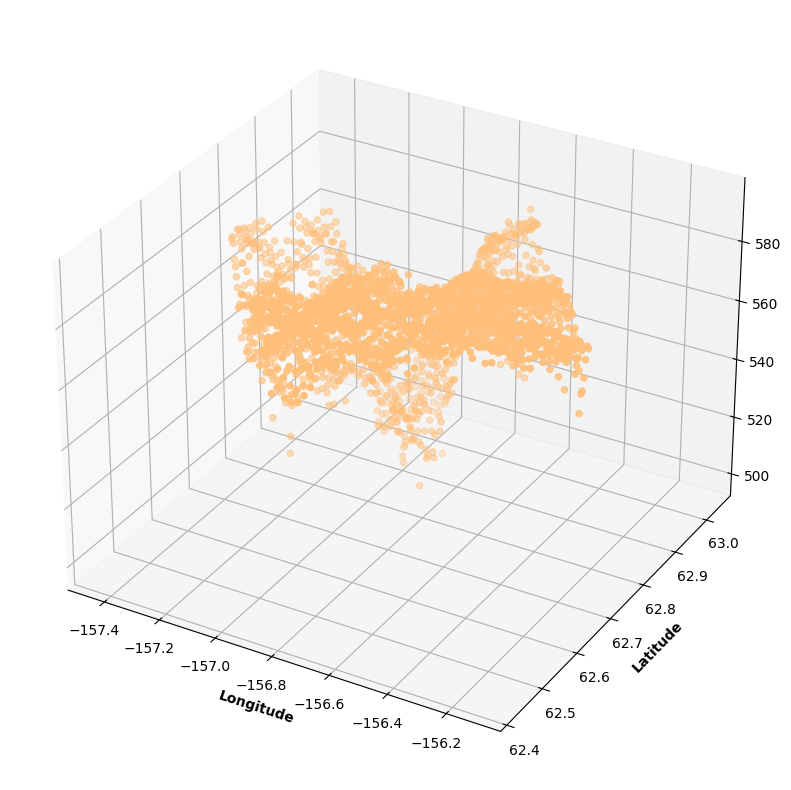}&
    			\includegraphics [width=0.12\textwidth]{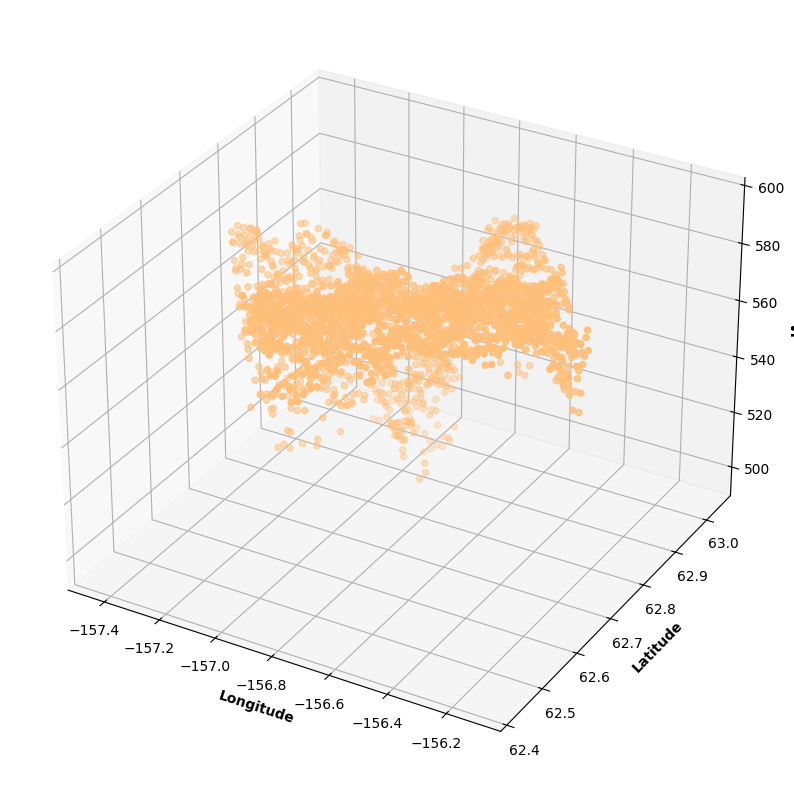}&
    			\includegraphics [width=0.12\textwidth]{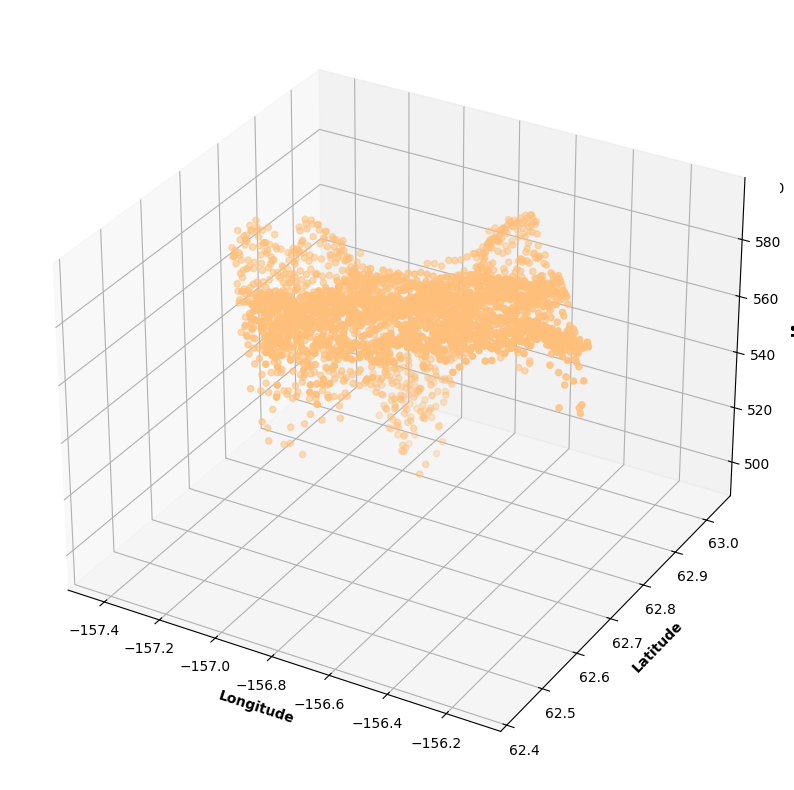}\\
    			Observed & DT & KNR & RF  & SIREN & S-INR & Original\\
    		\end{tabular}
    	\end{center}
    	\caption{From top to down respectively list the results of weather data completion by different methods on Prcp, SWE, and VP for the weather data located at ($63^{\circ}$N, $157^{\circ}$W) with the random sampling rate of 0.25.}
    	\label{figweather}
    \end{figure*}	
	\subsubsection{Weather Date Completion Results.}
	The results on weather data completion are shown in Table \ref{Multidatatable} and Fig. \ref{figweather}. Compared with other INR methods, our proposed S-INR achieves superior performance in quantitative results in terms of NRMSE and R-square in Table \ref{Multidatatable}. Visually, we can observe that the results of the proposed method are generally closest to the true values in Fig. \ref{figweather}. In contrast, other regression methods struggle to recover the complex structure of weather data, resulting in their results being quite different from the original realistic results. The superior performance of our method can be attributed to the elaborately designed components in S-INR to exploit the semantic information within and across generalized superpixels.


	\subsection{Discussions}
	\subsubsection{The Role of Generalized Superpixels.} 
	To better understand the roles of basic units, we compare two categories of basic units (i.e., pixels and generalized superpixels) based INR, namely the traditional pixel-based INR \cite{sitzmann2020implicit} and the superpixel-based INR. For fairness, we adopt the same INR architecture\footnote{Specifically, the superpixel-based INR is based on using $K$ INRs to represent $K$ superpixels, where $K$ is the number of generalized superpixels.} that uses MLP with the sinusoidal activation function to map the coordinates to corresponding values. Both model parameter scales are kept on the same scale by adjusting certain model hyperparameters. 
	
	\begin{figure*}[htbp!]
		\tiny
		\setlength{\tabcolsep}{0.9pt}
		\begin{center}
			\begin{tabular}{cccc}
				\includegraphics [width=0.2\textwidth]{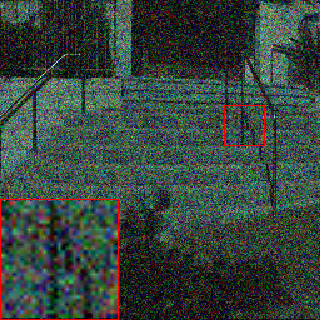}&
				\includegraphics [width=0.2\textwidth]{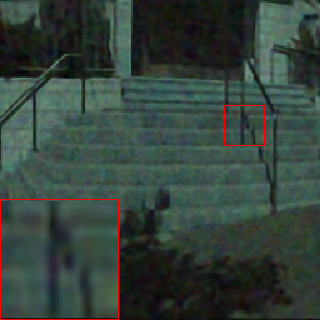}&
				\includegraphics [width=0.2\textwidth]{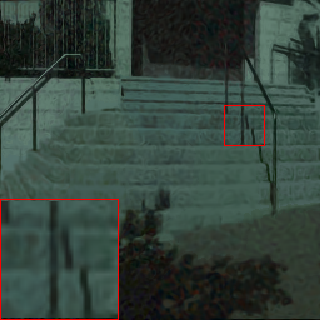}&
				\includegraphics [width=0.2\textwidth]{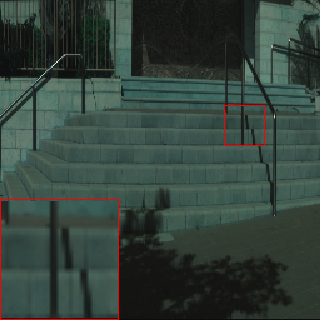}\\
				Observed & Pixel-based INR& Superpixel-based INR & Original \\
				PSNR: 13.977 dB & PSNR: 29.348 dB & \textbf{PSNR: 31.475 dB} & PSNR: Inf \\
			\end{tabular}
		\end{center}
		\caption{The results of image denoising by pixel-based INR\cite{sitzmann2020implicit} and superpixel-based INR on MSI $\it{Lehavim}$ in Case1. When utilizing our proposed generalized superpixels as basic units, INR can restore finer details and remove more noise by exploiting the inherent semantic information of the data. The \textbf{best} values are highlighted.}
		\label{fig:ablation1}
	\end{figure*}
	
			\begin{table}[htbp!]
		\centering
		\caption{Quantitative results of image denoising on $\it{Lehavim}$ in Case2. The \textbf{best} values are highlighted.}
		\resizebox{0.95\textwidth}{!}{
			\begin{tabular}{ccccccc}
				\toprule
				\multicolumn{3}{c}{\textbf{Components}} & \multicolumn{2}{c}{$\it{Lehavim}$} & \multirow{2}{*}{Parameters (Mb)} \\
				\cmidrule(lr){1-3} \cmidrule(lr){4-5}
				Generalized Superpixels & Exclusive MLPs & Shared Dictionary Matrix & PSNR & SSIM &  \\
				\midrule
				\XSolidBrush & \XSolidBrush & \XSolidBrush & 29.348 & 0.745 & 0.786 \\
				\Checkmark & \XSolidBrush & \XSolidBrush & 31.475 & 0.817 & \textbf{0.741} \\
				\Checkmark & \XSolidBrush & \Checkmark & 31.997 & 0.818 & 0.797 \\
				\Checkmark & \Checkmark & \XSolidBrush & 31.012 & 0.787 & 0.902 \\
				\Checkmark & \Checkmark & \Checkmark & \textbf{32.394} & \textbf{0.831} & 0.995 \\
				\bottomrule
		\end{tabular}}
		\label{tab:ablation1}
	\end{table}

	We conduct quantitative and qualitative comparisons between the pixel-based INR and the superpixel-based INR, with results presented in Fig. \ref{fig:ablation1}. We can observe that superpixel-based INR performs better than pixel-based INR as evidenced by higher PSNR. Visually, we can observe that superpixel-based INR preserves finer details with fewer artifacts than pixel-based INR, e.g., we can find in the zoomed region that superpixel-based INR can represent more intricate details, while its background contains less noise. The superior performance of superpixel-based INR is attributed to utilizing our proposed basic unit, namely generalized superpixel, which encode rich inherent semantic information of the data.

	

	\subsubsection{Ablation Experiment.}
	To understand the roles of S-INR's key components, we perform an ablation experiment on the image denoising task. The important components include the use of generalized superpixels as basic units (generalized superpixels), exclusive attention-based MLPs (exclusive MLPs), and a shared dictionary matrix.

    Table \ref{tab:ablation1} shows the recovery results of quantitative comparisons with different components in S-INR. We can observe that the three components in S-INR all contribute to the superior performance of S-INR. The results of S-INR without the generalized superpixels as basic units, exclusive attention-based MLPs, and the shared dictionary matrix show the performance degradation of approximately 3.0dB, 0.4dB, and 1.4dB in terms of PSNR, respectively, compared to the performance of our proposed S-INR. This demonstrates that all three components are important to S-INR, and by working together, they can effectively exploit the inherent semantic information of the data to achieve state-of-the-art performance.

	\section{Conclusion}
	In this work, we first suggested utilizing generalized superpixels instead of pixels as alternative basic units of INR for multi-dimensional data that encode rich semantic information. Then, to fully exploit the semantic information within and across generalized superpixels, we proposed a novel superpixel-informed implicit neural representation (termed as S-INR), which includes two elaborately designed modules, namely the exclusive attention-based MLPs and a shared dictionary matrix. Extensive experiments have validated consistent improvements in our proposed S-INR over state-of-the-art INR methods in terms of quantitative metrics and visual quality. We believe that S-INR is a potential tool for more versatile multi-dimensional applications.
	
	
	%
	%
	
	\noindent\textbf{Acknowledgments:} This research is supported by NSFC (No. 12371456, 12171-072, 62131005, 62306248), Sichuan Science and Technology Program (No. 2024NSFJQ0038, 2023ZYD0007), and National Key Research and Development Program of China (No. 2020YFA0714001).
	
	\bibliographystyle{splncs04}
	\bibliography{egbib}

\begin{thebibliography}{10}
\providecommand{\url}[1]{\texttt{#1}}
\providecommand{\urlprefix}{URL }
\providecommand{\doi}[1]{https://doi.org/#1}

\bibitem{knr}
Altman, N.S.: An introduction to kernel and nearest-neighbor nonparametric
  regression. The American Statistician  \textbf{46}(3),  175--185 (1992)

\bibitem{kmeansj}
Arthur, D., Vassilvitskii, S.: k-means++: the advantages of careful seeding.
  In: Proceedings of the Eighteenth Annual ACM-SIAM Symposium on Discrete
  Algorithms. pp. 1027--1035 (2007)

\bibitem{point2}
Ben-Shabat, Y., Koneputugodage, C.H., Gould, S.: Digs: Divergence guided shape
  implicit neural representation for unoriented point clouds. In: Proceedings
  of the IEEE/CVF Conference on Computer Vision and Pattern Recognition (CVPR).
  pp. 19323--19332 (June 2022)

\bibitem{wa4}
Cervantes, P., Sekikawa, Y., Sato, I., Shinoda, K.: Implicit neural
  representations for variable length human motion generation. In: Proceedings
  of the European Conference Computer Vision (ECCV). pp. 356--372 (2022)

\bibitem{vedio2}
Chen, H., He, B., Wang, H., Ren, Y., Lim, S.N., Shrivastava, A.: Nerv: Neural
  representations for videos. In: Proceedings of the International Conference
  on Neural Information Processing Systems (NeurIPS). vol.~34, pp. 21557--21568
  (2021)

\bibitem{image4}
Chen, Y., Liu, S., Wang, X.: Learning continuous image representation with
  local implicit image function. In: Proceedings of the IEEE/CVF Conference on
  Computer Vision and Pattern Recognition (CVPR). pp. 8628--8638 (June 2021)

\bibitem{spacetime}
Davis, J., Nehab, D., Ramamoorthi, R., Rusinkiewicz, S.: Spacetime stereo: a
  unifying framework for depth from triangulation. IEEE Transactions on Pattern
  Analysis and Machine Intelligence  \textbf{27}(2),  296--302 (2005)

\bibitem{mfn}
Fathony, R., Sahu, A.K., Willmott, D., Kolter, J.Z.: Multiplicative filter
  networks. In: Proceedings of the International Conference on Learning
  Representations (ICLR) (2021)

\bibitem{point1}
Fujiwara, K., Hashimoto, T.: Neural implicit embedding for point cloud
  analysis. In: Proceedings of the IEEE/CVF Conference on Computer Vision and
  Pattern Recognition (CVPR). pp. 11734--11743 (June 2020)

\bibitem{t1}
Henzler, P., Mitra, N.J., Ritschel, T.: Learning a neural \text{3D} texture
  space from \text{2D} exemplars. In: Proceedings of the IEEE/CVF Conference on
  Computer Vision and Pattern Recognition (CVPR). pp. 8353--8361 (June 2020)

\bibitem{rf}
Ho, T.K.: Random decision forests. In: Proceedings of 3rd International
  Conference on Document Analysis and Recognition (ICDAR). vol.~1, pp. 278--282
  (1995)

\bibitem{Hofherr_2023_WACV}
Hofherr, F., Koestler, L., Bernard, F., Cremers, D.: Neural implicit
  representations for physical parameter inference from a single video. In:
  Proceedings of the IEEE/CVF Winter Conference on Applications of Computer
  Vision (WACV). pp. 2093--2103 (January 2023)

\bibitem{wa1}
Hong, S., Nam, J., Cho, S., Hong, S., Jeon, S., Min, D., Kim, S.: Neural
  matching fields: Implicit representation of matching fields for visual
  correspondence. In: Proceedings of the International Conference on Neural
  Information Processing Systems (NeurIPS). vol.~35, pp. 13512--13526 (2022)

\bibitem{se}
Hu, J., Shen, L., Albanie, S., Sun, G., Wu, E.: \text{Squeeze-and-Excitation}
  networks. IEEE Transactions on Pattern Analysis and Machine Intelligence
  \textbf{42}(8),  2011--2023 (August 2020)

\bibitem{dt}
Hunt, E.B., Marin, J., Stone, P.J.: Experiments in induction.  (1966)

\bibitem{psnr}
Huynh-Thu, Q., Ghanbari, M.: Scope of validity of psnr in image/video quality
  assessment. Electronics Letters  \textbf{44},  800--801 (02 2008)

\bibitem{nips_mlp_high_fre}
Jacot, A., Gabriel, F., Hongler, C.: Neural tangent kernel: Convergence and
  generalization in neural networks. In: Proceedings of the International
  Conference on Neural Information Processing Systems (NeurIPS). vol.~31, p.~6
  (2018)

\bibitem{n2}
Jayasundara, V., Agrawal, A., Heron, N., Shrivastava, A., Davis, L.S.:
  \text{FlexNeRF}: Photorealistic free-viewpoint rendering of moving humans
  from sparse views. In: Proceedings of the IEEE/CVF Conference on Computer
  Vision and Pattern Recognition (CVPR). pp. 21118--21127 (June 2023)

\bibitem{n3}
Kaneko, T.: \text{MIMO-NeRF}: Fast neural rendering with multi-input
  multi-output neural radiance fields. In: Proceedings of the IEEE/CVF
  International Conference on Computer Vision (ICCV). pp. 3273--3283 (October
  2023)

\bibitem{auto1}
Kazerouni, A., Azad, R., Hosseini, A., Merhof, D., Bagci, U.: Incode: Implicit
  neural conditioning with prior knowledge embeddings. In: Proceedings of the
  IEEE/CVF Winter Conference on Applications of Computer Vision (WACV). pp.
  1298--1307 (January 2024)

\bibitem{images2}
Khan, M.O., Fang, Y.: Implicit neural representations for medical imaging
  segmentation. In: Proceedings of the Medical Image Computing and Computer
  Assisted Intervention (MICCAI). pp. 433--443 (2022)

\bibitem{adam}
Kingma, D., Ba, J.: Adam: A method for stochastic optimization. Proceedings of
  the International Conference on Learning Representations (ICLR) p.~13 (May
  2015)

\bibitem{bacon}
Lindell, D.B., Van~Veen, D., Park, J.J., Wetzstein, G.: \text{BACON}:
  Band-limited coordinate networks for multiscale scene representation. In:
  Proceedings of the IEEE/CVF Conference on Computer Vision and Pattern
  Recognition (CVPR). pp. 16252--16262 (June 2022)

\bibitem{vedio1}
Lu, Y., Wang, Z., Liu, M., Wang, H., Wang, L.: Learning spatial-temporal
  implicit neural representations for event-guided video super-resolution. In:
  Proceedings of the IEEE/CVF Conference on Computer Vision and Pattern
  Recognition (CVPR). pp. 1557--1567 (June 2023)

\bibitem{image3}
Ma, C., Yu, P., Lu, J., Zhou, J.: Recovering realistic details for
  magnification-arbitrary image super-resolution. IEEE Transactions on Image
  Processing  \textbf{31},  3669--3683 (2022)

\bibitem{Mai_2022_CVPR}
Mai, L., Liu, F.: Motion-adjustable neural implicit video representation. In:
  Proceedings of the IEEE/CVF Conference on Computer Vision and Pattern
  Recognition (CVPR). pp. 10738--10747 (June 2022)

\bibitem{acorn}
Martel, J.N.P., Lindell, D.B., Lin, C.Z., Chan, E.R., Monteiro, M., Wetzstein,
  G.: Acorn: adaptive coordinate networks for neural scene representation. ACM
  Transactions on Graphics  \textbf{40}(4) (July 2021)

\bibitem{nerf}
Mildenhall, B., Srinivasan, P.P., Tancik, M., Barron, J.T., Ramamoorthi, R.,
  Ng, R.: \text{NeRF}: Representing scenes as neural radiance fields for view
  synthesis. In: Proceedings of the European Conference Computer Vision (ECCV).
  pp. 405--421 (2020)

\bibitem{t2}
Oechsle, M., Mescheder, L., Niemeyer, M., Strauss, T., Geiger, A.: Texture
  fields: Learning texture representations in function space. In: Proceedings
  of the IEEE/CVF International Conference on Computer Vision (ICCV). pp.
  4531--4540 (October 2019)

\bibitem{pmlr_mlp_high_fre}
Rahaman, N., Baratin, A., Arpit, D., Draxler, F., Lin, M., Hamprecht, F.,
  Bengio, Y., Courville, A.: On the spectral bias of neural networks. In:
  Proceedings of the 36th International Conference on Machine Learning (ICML).
  vol.~97, pp. 5301--5310 (June 2019)

\bibitem{eccv_activation}
Ramasinghe, S., Lucey, S.: Beyond \text{Periodicity}: Towards a unifying
  framework for activations in coordinate-mlps. In: Proceedings of the European
  Conference Computer Vision (ECCV). pp. 142--158 (2022)

\bibitem{n1}
Rebain, D., Jiang, W., Yazdani, S., Li, K., Yi, K.M., Tagliasacchi, A.:
  \text{DeRF}: Decomposed radiance fields. In: Proceedings of the IEEE/CVF
  Conference on Computer Vision and Pattern Recognition (CVPR). pp.
  14153--14161 (June 2021)

\bibitem{inverse2}
Reed, A.W., Kim, H., Anirudh, R., Mohan, K.A., Champley, K., Kang, J.,
  Jayasuriya, S.: Dynamic \text{CT} reconstruction from limited views with
  implicit neural representations and parametric motion fields. In: Proceedings
  of the IEEE/CVF International Conference on Computer Vision (ICCV). pp.
  2258--2268 (October 2021)

\bibitem{kilo}
Reiser, C., Peng, S., Liao, Y., Geiger, A.: \text{KiloNeRF}: Speeding up neural
  radiance fields with thousands of tiny mlps. In: Proceedings of the IEEE/CVF
  International Conference on Computer Vision (ICCV). pp. 14335--14345 (October
  2021)

\bibitem{WIRE}
Saragadam, V., LeJeune, D., Tan, J., Balakrishnan, G., Veeraraghavan, A.,
  Baraniuk, R.G.: \text{WIRE}: Wavelet implicit neural representations. In:
  Proceedings of the IEEE/CVF Conference on Computer Vision and Pattern
  Recognition (CVPR). pp. 18507--18516 (June 2023)

\bibitem{miner}
Saragadam, V., Tan, J., Balakrishnan, G., Baraniuk, R.G., Veeraraghavan, A.:
  \text{MINER}: Multiscale implicit neural representations. In: Proceedings of
  the European Conference Computer Vision (ECCV). pp. 318--333 (2022)

\bibitem{serrano2024hosc}
Serrano, D., Szymkowiak, J., Musialski, P.: Hosc: A periodic activation
  function for preserving sharp features in implicit neural representations
  (2024)

\bibitem{shen2023trident}
Shen, Z., Cheng, Y., Chan, R.H., Liò, P., Schönlieb, C.B., Aviles-Rivero,
  A.I.: Trident: The nonlinear trilogy for implicit neural representations
  (2023)

\bibitem{sitzmann2020implicit}
Sitzmann, V., Martel, J., Bergman, A., Lindell, D., Wetzstein, G.: Implicit
  neural representations with periodic activation functions. Proceedings of the
  International Conference on Neural Information Processing Systems (NeurIPS)
  \textbf{33},  7462--7473 (2020)

\bibitem{auto2}
Su, K., Chen, M., Shlizerman, E.: Inras: Implicit neural representation for
  audio scenes. In: Proceedings of the International Conference on Neural
  Information Processing Systems (NeurIPS). vol.~35, pp. 8144--8158 (2022)

\bibitem{inverse1}
Sun, Y., Liu, J., Xie, M., Wohlberg, B., Kamilov, U.S.: \text{CoIL}:
  Coordinate-based internal learning for tomographic imaging. IEEE Transactions
  on Computational Imaging  \textbf{7},  1400--1412 (2021)

\bibitem{tancik2020fourfeat}
Tancik, M., Srinivasan, P., Mildenhall, B., Fridovich-Keil, S., Raghavan, N.,
  Singhal, U., Ramamoorthi, R., Barron, J., Ng, R.: Fourier features let
  networks learn high frequency functions in low dimensional domains. In:
  Proceedings of the International Conference on Neural Information Processing
  Systems (NeurIPS). vol.~33, pp. 7537--7547 (2020)

\bibitem{dip}
Ulyanov, D., Vedaldi, A., Lempitsky, V.: Deep image prior. In: Proceedings of
  the IEEE Conference on Computer Vision and Pattern Recognition (CVPR). pp.
  9446--9454 (June 2018)

\bibitem{images3}
Umpler, Y., Postels, J., Yang, R., Gool, L.V., Tombari, F.: Implicit neural
  representations for image compression. In: Proceedings of the European
  Conference Computer Vision (ECCV). pp. 74--91 (2022)

\bibitem{ssim}
Wang, Z., Bovik, A., Sheikh, H., Simoncelli, E.: Image quality assessment: from
  error visibility to structural similarity. IEEE Transactions on Image
  Processing  \textbf{13}(4),  600--612 (2004)

\bibitem{t3}
Xiang, F., Xu, Z., Hašan, M., Hold-Geoffroy, Y., Sunkavalli, K., Su, H.:
  \text{NeuTex}: Neural texture mapping for volumetric neural rendering. In:
  Proceedings of the IEEE/CVF Conference on Computer Vision and Pattern
  Recognition (CVPR). pp. 7119--7128 (June 2021)

\bibitem{inpainting1}
Xu, D., Wang, P., Jiang, Y., Fan, Z., Wang, Z.: Signal processing for implicit
  neural representations. In: Proceedings of the International Conference on
  Neural Information Processing Systems (NeurIPS). vol.~35, pp. 13404--13418
  (2022)

\bibitem{images1}
Xu, R., Yao, M., Chen, C., Wang, L., Xiong, Z.: Continuous spectral
  reconstruction from \text{RGB} images via implicit neural representation.
  In: Proceedings of the European Conference Computer Vision (ECCV). pp. 78--94
  (2023)

\bibitem{inpainting2}
Xu, W., Jiao, J.: Revisiting implicit neural representations in low-level
  vision. In: International Conference on Learning Representations Workshop
  (2023)

\bibitem{simulation}
Yang, L., Kim, B., Zoss, G., G\"{o}zc\"{u}, B., Gross, M., Solenthaler, B.:
  Implicit neural representation for physics-driven actuated soft bodies. ACM
  Transactions on Graphics  \textbf{41}(4) (July 2022)

\bibitem{image1}
Yang, S., Ding, M., Wu, Y., Li, Z., Zhang, J.: Implicit neural representation
  for cooperative low-light image enhancement. In: Proceedings of the IEEE/CVF
  International Conference on Computer Vision (ICCV). pp. 12918--12927 (October
  2023)

\bibitem{9991174}
Zhang, K., Zhu, D., Min, X., Zhai, G.: Implicit neural representation learning
  for hyperspectral image super-resolution. IEEE Transactions on Geoscience and
  Remote Sensing  \textbf{61},  1--12 (2023)

\bibitem{vedio3}
Zhao, Q., Asif, M.S., Ma, Z.: Dnerv: Modeling inherent dynamics via difference
  neural representation for videos. In: Proceedings of the IEEE/CVF Conference
  on Computer Vision and Pattern Recognition (CVPR). pp. 2031--2040 (June 2023)

\end{thebibliography}
\end{document}